\RequirePackage{fix-cm}
\documentclass[10pt,letterpaper]{paper}
\usepackage[authoryear,round]{natbib}
\usepackage[all]{hypcap}
\definecolor{title_blue}{HTML}{204899}
\definecolor{cite_blue}{HTML}{044DC1}
\definecolor{cite_purple}{HTML}{7406A7}
\hypersetup{
  colorlinks=true,
  citecolor=cite_blue,
  linkcolor=cite_purple,
  urlcolor=cite_purple,
}
\usepackage{iftex}
\ifPDFTeX
  \usepackage{CJKutf8}
  \newenvironment{ExpMathDraft}{\begin{CJK*}{UTF8}{gbsn}}{\end{CJK*}}
\else
  \usepackage{xeCJK}
\fi
\usepackage{inconsolata}

\usepackage{amsmath,amsfonts,bm}

\def\eqref#1{equation~\ref{#1}}

\def\1{\bm{1}}

\DeclareMathAlphabet{\mathsfit}{\encodingdefault}{\sfdefault}{m}{sl}
\SetMathAlphabet{\mathsfit}{bold}{\encodingdefault}{\sfdefault}{bx}{n}

\newcommand{\softmax}{\mathrm{softmax}}

\usepackage{hyperref}
\usepackage{url}
\usepackage{graphicx}
\usepackage{booktabs}
\usepackage{multirow}
\usepackage{xcolor}
\usepackage{array,tabularx}
\usepackage{listings}
\usepackage{flafter}
\usepackage{wrapfig}
\definecolor{ExpOurs}{HTML}{EBF3FF}
\definecolor{ExpReference}{HTML}{F2F3F3}
\definecolor{ExpBand}{HTML}{F5F6F6}

\newlength{\ExpMethodWidth}
\newlength{\ExpNumWidth}

\newcommand{\ExpLongBenchTaskCount}{15}
\newcommand{\ExpLongBenchSampleCount}{3,550}

\newcommand{\CompactParagraphs}{}

\title{The Gating Mechanism of KV Cache Eviction}

\author[1,*]{Hao Gu}
\author[2,*]{Xintong Yang}
\author[3]{Zhiwei Li}
\author[4]{Lei Zhu}
\author[1]{Jian Yang}
\author[1]{Qiyuan Zhu}
\author[1]{Dezhi Li}
\author[5]{Jingyuan Zhang}
\author[1]{Zhizhuo KOU}
\author[6]{Hao Wang}
\author[1]{Bei Liu}
\author[7]{Jiacheng Liu}
\author[1,\textdagger]{Sirui Han}

\affil[1]{The Hong Kong University of Science and Technology}
\affil[2]{New York University}
\affil[3]{The Hong Kong University of Science and Technology (Guangzhou)}
\affil[4]{Tencent HunYuan}
\affil[5]{Georgia Institute of Technology}
\affil[6]{City University of Hong Kong}
\affil[7]{Shanghai Jiaotong University}

\correspondingauthor{\textsuperscript{*}Equal contribution. \textsuperscript{\textdagger}Corresponding author: Sirui Han. Contact: \email{marcusguhao@gmail.com}.}
\renewcommand{\paperresources}{%
\href{https://github.com/Chooovy/IndexMem}{Code: IndexMem++}\qquad
\href{https://huggingface.co/collections/marcusguhao/indexmem}{Checkpoints: Hugging Face}}
\reportnumber{}

\begin{abstract}
\textbf{Abstract:}
KV cache eviction must decide which entries carry information that the rest of the cache can no longer supply. A natural way to estimate this is a learnable gate: give each cached entry a gate on its attention weight and train the gate on the model's own objective. The gate belongs inside the normalizing softmax, where suppressing one entry hands its mass to the others rather than discarding it. That same normalizer absorbs any per-query constant, so a constant gate cancels and reproduces the frozen model exactly. Trained on the language-modeling loss the gate stays near a constant score while its loss falls throughout: optimization succeeds, but a frozen backbone was fit for dense attention, so its loss prefers the no-op, and a near-constant score ranks nothing. The no-op is not a floor the gate improves from but what its objective asks for. A second obstacle survives once the gate moves, since the same gradient charges the gate for errors the frozen model already makes, which no reweighting of positions separates out. Much of the machinery in learnable eviction, from sparsity penalties to relaxations of the discrete decision, works around the first obstacle rather than removing it; both are removed by putting dense attention beyond the gate's reach and making the dense model's predictions its target. Under these two conditions capacity is never the missing ingredient: neither a larger scorer, nor explicit access to the past, nor a trainable backbone helps. The same holds for what eviction unavoidably discards, which is recovered by parameter-free centroids read inside the same softmax far better than by any memory read outside it, since the mass they must return is a share of that normalizer and available nowhere else. At 75\% KV cache compression, this gate reaches 90.14 on RULER at 8K and 46.03 on LongBench with Qwen3-8B, 3.72 and 1.36 points below full-cache inference.
\end{abstract}

\begin{document}
\maketitle

\addtocontents{toc}{\protect\setcounter{tocdepth}{-5}}

\section{Introduction}
\label{sec:intro}
Large language models are evolving from chatbot assistants~\citep{touvron2023llama} into long-horizon agents~\citep{yao2023react}. These agentic workloads naturally induce long interaction histories, making efficient long-context inference essential.
During autoregressive decoding, the KV cache is increasingly the main bottleneck. As the cache grows linearly with sequence length, KV-cache storage and data movement become the dominant systems bottlenecks~\citep{kwon2023pagedattention}. Reducing this footprint has therefore become a central objective in efficient LLM inference. Existing approaches compress KV cache along several axes: quantization reduces precision~\citep{liu2024kivi}, cross-layer sharing reduces number of layer-specific KV~\citep{sun2024yoco}, and latent representations reduce KV dimension~\citep{deepseekai2024deepseekv2}. These approaches are complementary and can be combined. 
In this work, we focus on KV eviction because it not only reduces memory consumption but also shortens the effective sequence length of attention, thereby reducing attention computation.
KV eviction is possible because information becomes redundant across the cache. Through causal self-attention, each hidden state aggregates preceding context, allowing later KV entries to carry information originally introduced by earlier tokens. This layer-wise propagation explains why sliding-window attention can access information beyond its explicit window~\citep{fu2025swat}.

However, this propagation is incomplete: isolated facts such as passwords or retrieval keys may remain unique to their original entries. Effective eviction must therefore distinguish two kinds of entries: those whose information can be recovered from the rest of the cache, and those that carry the only copy of what they contribute.
Sparse attention and KV eviction appear similar because both select a subset of KV entries, but their objectives differ. Sparse attention selects entries relevant to the current query; eviction must identify entries that are redundant regardless of future queries.
Existing eviction methods however often treat relevance as redundancy. H$_2$O and SnapKV use attention from previously observed queries~\citep{zhang2023h2o,li2024snapkv}. Relevance can serve as a partial proxy for redundancy, but does not measure it directly.

We instead define redundancy relative to the retained cache: an entry is redundant if its information can be recovered from the other KV entries that remain. Hand-crafted measures such as cosine similarity~\citep{keydiff,rkv} capture only limited aspects of this relationship. We therefore learn a lightweight, query-independent gate that maps each contextualized hidden state to a retention score for each KV head. And where that gate is attached is not a free choice. Attention weights are shares of one softmax normalizer, so only a gate inside that normalizer hands a suppressed entry's mass to the remaining entries, which is what removal means.

Naively training this gate does not work: it learns nearly identical retention scores for all tokens, reproducing dense attention without learning which entries can be evicted. The reason is structural and it comes from the same normalizer that gives the gate its meaning. Because a softmax absorbs any per-query constant, a constant score cancels inside it and is a no-op, and a frozen backbone's loss prefers exactly that. The gate must therefore be denied dense attention and asked to match the dense model's predictions instead of the next token. The same invariance also bounds what a score can say: it ranks entries within a head but cannot compare heads, so the division of budget among them must be settled outside the score.

\begin{wrapfigure}{r}{0.33\linewidth}
\vspace{-15pt}
\centering
\includegraphics[width=\linewidth]{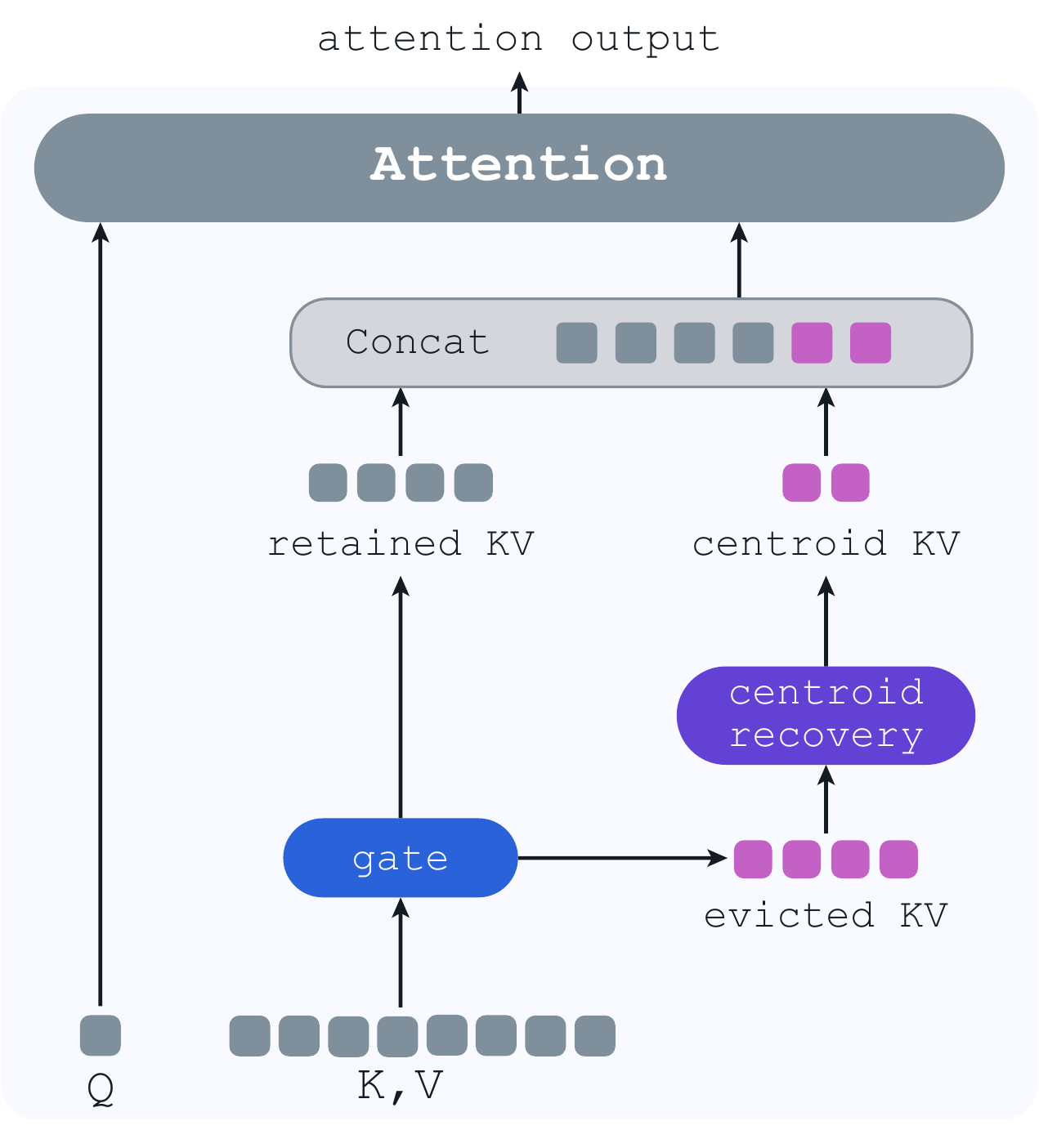}
\vspace{-15pt}
\caption{\small The two branches of IndexMem++: evict and recovery both inside one softmax.}
\label{fig:overview}
\vspace{-8pt}
\end{wrapfigure}
Even accurate retention scores cannot make hard eviction lossless: retained entries may preserve part of an evicted entry's information, but not all of it. We find that discarded entries still receive substantial attention mass, motivating a compact mechanism for recovering their residual contribution. Our key finding is that the readout of this compressed information matters more than the capacity of the memory that stores it, and the reason is the normalizer again. Recovering the missing contribution requires both its direction and the fraction of attention mass it holds, and that fraction is a ratio of softmax denominators: it is available exactly where the gate was, and only estimable indirectly anywhere else. A memory read outside attention must learn that calibration in output space and does not improve as it grows, whereas parameter-free centroids of the evicted keys and values, inserted into the cache and weighted by the same softmax as the retained entries, recover it far more closely. Removal and restoration are thus the same statement about where the normalizer applies. We instantiate these findings in IndexMem++, which follows a two-branch design~\citep{indexmem}, summarized in Figure~\ref{fig:overview}: a selection branch evicts redundant KV entries, and a recovery branch summarizes what was evicted, both reading through one softmax. Experiments on Qwen3-4B and Qwen3-8B show consistent gains across long-context understanding and mathematical reasoning. At 75\% compression, IndexMem++ reaches 90.14 on RULER with Qwen3-8B and 46.03 on LongBench, within 1.36 points of Full-KV, and it retains 0.807 pass@1 on Math500 with a 512-entry cache. Section~\ref{sec:exp} reports the full comparison.

\paragraph{Contributions.}
Our contributions are threefold:
{\setlength{\leftmargini}{1.3em}
\begin{itemize}
\item We distinguish redundancy from query relevance, formulate KV eviction as query-independent redundancy estimation, and unify learnable gate methods under a common design space.
\item We show that the same normalizer is why naive gate training fails: it absorbs a per-query constant, so dense attention is always reachable and is what a frozen backbone's loss prefers. Learning a usable score requires denying the gate that solution, by exempting the entries that must be retained anyway and making the rest compete for a fixed mass, and supervising it against the dense model rather than the next token. The one thing the score cannot express, the division of budget across heads, follows from the same invariance and is set by retained attention mass.
\item We show that readout matters more than memory capacity for recovering evicted information, because the missing mass fraction is a ratio of softmax denominators, and introduce content-clustered centroid KV tokens that re-enter the attention softmax at a matched cache budget.
\end{itemize}}

\section{Attention mass and the eviction gate}
\label{sec:prelim}

\paragraph{Attention and KV eviction.}
Writing $h_t$ for the hidden state of token $t$ and $d$ for the head dimension, self-attention projects $h_t$ into a query, key, and value, and token $t$ attends to all preceding tokens:
\begin{equation}
    o_t=\sum_{j\leq t}a_{tj}v_j,
    \qquad
    a_{tj}=\frac{\exp(\ell_{tj})}{Z_t},
    \qquad
    \ell_{tj}=\frac{q_tk_j^{\top}}{\sqrt{d}},
    \qquad
    Z_t=\sum_{\ell\leq t}\exp(\ell_{t\ell}).
\end{equation}
The weights $a_{tj}$ are shares of the single normalizer $Z_t$ and sum to one; we call $a_{tj}$ the \emph{attention mass} that query $t$ assigns to entry $j$, and it is the quantity the rest of this paper is about. The keys and values are cached and reused at every decoding step, so the cache grows linearly with context length. Under a budget of $B$ entries, an eviction policy scores each entry and keeps the current top-$B$, selecting at every step among the surviving entries and the new one; everything outside the top-$B$ is physically removed and never reconsidered.

\paragraph{From hard eviction to a differentiable gate.}
A learned gate or router can score entries for sparse top-$B$ selection, but learning which entries to select requires a training signal through a discrete operation: the selected indices are piecewise constant in the scores. Sparse mixture-of-experts models preserve a gradient path by multiplying selected expert outputs by their continuous routing weights~\citep{shazeer2017moe}. DeepSeek Sparse Attention (DSA)~\citep{dsa} takes another route: it trains its indexer to match the main model's attention distribution, initially under dense attention, rather than differentiating through the top-$k$ choice.

KV eviction, however, lacks a suitable \emph{eviction oracle}. An attention weight measures how much one query reads an entry; eviction must decide whether permanently removing it will harm later predictions given the information that remains. KVzap~\citep{kvzap} and Fast KVzip~\citep{fastkvzip} learn KVzip-style reconstruction scores from hidden states, avoiding backpropagation through the eviction decision. These targets measure importance during context reconstruction, not the prediction error caused by removing an entry from a budgeted cache. Even exact prediction of the proxy would therefore leave the mismatch with the eviction objective unresolved.

\begin{figure}[!htbp]
\centering
\includegraphics[width=\linewidth]{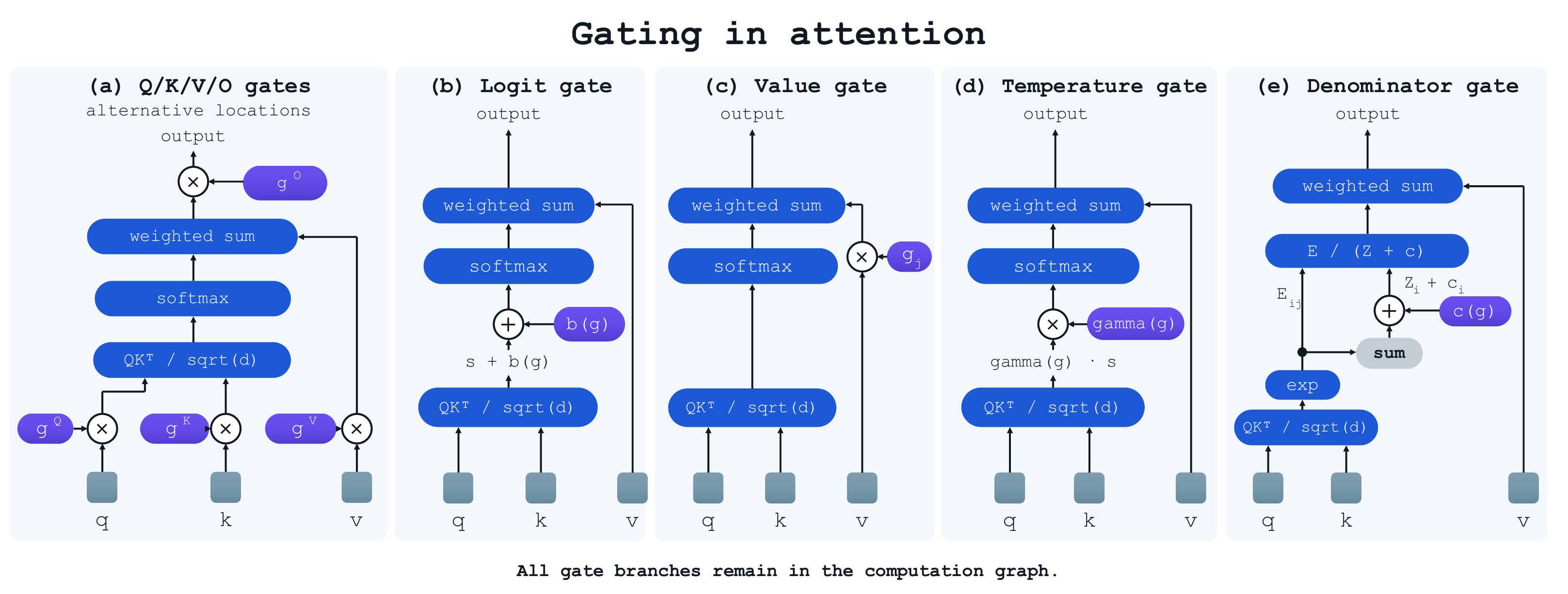}
\caption{\textbf{Differentiable gates can act at different points in attention.} (a) Gates on queries, keys, values, or outputs; (b) an additive logit bias; (c) a gate on the value branch; (d) a multiplicative logit scale; and (e) a modification of the denominator alone. All keep the gate in the computation graph, but they have different effects on attention mass. A value gate suppresses a contribution without renormalizing the other weights, whereas a per-key logit bias $\log g_j$ redistributes mass and recovers attention over the retained set in the binary limit.}
\label{fig:gate-in-attention}
\end{figure}

We instead train retention scores through the model's predictions, replacing the binary keep-or-drop decision during training with a continuous per-entry gate $g_j\geq 0$. A differentiable path is obtained by inserting the gate's output into attention's computation graph. This admits many choices: gates on queries, keys, values, or outputs; additive logit biases; temperature or denominator gates; and even nested softmax constructions. Figure~\ref{fig:gate-in-attention} illustrates representative placements. Differentiability alone does not determine which placement approximates eviction, because removing an entry also removes its key from the normalizer and redistributes its attention mass. We therefore focus on the distinction between a gate \emph{outside} the softmax, which scales a value contribution, and a gate \emph{inside} the softmax, which renormalizes attention over the gated entries.

\paragraph{After the softmax: suppressing a value.}
An outer gate multiplies each already-normalized attention weight, equivalently scaling the corresponding value:
\begin{equation*}
o_i^{\mathrm{outer}}=\sum_j a_{ij} g_j v_j.
\end{equation*}
Setting $g_j=0$ removes its value contribution, but its key still contributes to the original normalizer $Z_i$. The other weights stay unchanged, and the effective weights $a_{ij}g_j$ need not sum to one. This is attenuation of the retrieved message; it does not reproduce attention over the surviving cache.

\paragraph{Inside the softmax: removing a cache entry.}
We instead renormalize after applying the gates:
\begin{equation}
    a_{ij}^{(g)}=\frac{a_{ij}\,g_j}{\sum_{m}a_{im}\,g_m},
    \qquad
    o_i=\sum_{j=1}^{i}a_{ij}^{(g)}v_j.
    \label{eq:inner-gate}
\end{equation}
The denominator transfers the mass withdrawn from one entry to the others. For a binary gate with at least one retained entry, Equation~\ref{eq:inner-gate} is exactly attention over the retained set; positive continuous gates provide its differentiable relaxation. Since $\exp(\ell_{ij})g_j=\exp(\ell_{ij}+\log g_j)$, this is an additive bias on each key's attention logit, with $g_j\to 0$ recovering the hard mask $\log g_j\to-\infty$. The gate is independent of future query content, while its effect is measured relative to the other entries through the shared normalizer.

\paragraph{Other placements have different meanings.}
Query and key gates change the retrieval logits, whereas value and output gates scale the retrieved payload. Reducing a multiplicative logit scale pulls its logit towards zero, whose effect depends on the logit's sign under the current query; adding mass only to the denominator suppresses the row without providing a per-entry retention ranking. Appendix~\ref{sec:gating} gives the detailed comparison. Our construction specifically uses a per-key additive logit bias to relax cache removal.

\paragraph{The paradigm.}
The direct instantiation of this framework is
\begin{equation}
    \mathbf{a}_i=\operatorname{softmax}\!\left(
    \frac{q_iK_{\leq i}^{\top}}{\sqrt d}+\log\mathbf{g}_{\leq i}
    \right),
    \qquad
    s_j=f_\theta(h_j),
    \qquad
    \log g_j=s_j,
    \label{eq:paradigm}
\end{equation}
where $f_\theta$ maps the contextualized hidden state $h_j$ to a per-KV-head retention score. During training, the backbone is frozen and gradients through gated attention update $\theta$. During inference, the scores select which KV entries remain; attention over the retained entries does not apply the training-time soft gates. A uniform positive gate still cancels inside the softmax and recovers dense attention. Section~\ref{sec:method} explains why this makes direct training fail, how to train a useful ranking, and how the recovery branch of Figure~\ref{fig:overview} restores part of the discarded contribution through the same normalizer.

\section{Method}
\label{sec:method}
Equation~\ref{eq:paradigm} fixes the mechanism but not the design.
We begin from its most direct instantiation and find that it does not work at all, for a reason that is structural rather than a matter of tuning. The shared normalizer that gives the gate its meaning has a second consequence: it absorbs any per-query constant, so dense attention is always within the gate's reach. This section follows that consequence through, to what has to be taken away from the gate, what has to be given to it, and the one question a retention score is unable to answer. A pilot ablation on Qwen3-0.6B isolates each step.

\paragraph{Pilot setup.}
All ablations use Qwen3-0.6B on RULER at 8K with a cache budget of $2048$ entries, $25\%$ of the context. Every arm trains for 100 steps and differs from the reference in a single field.
Appendix~\ref{app:ablation} gives the full protocol, the per-task scores.

\begin{table}[!t]
\centering
\caption{
    \textbf{Each component of the paradigm, added one at a time.} Qwen3-0.6B, RULER 8K, cache budget 2048 of 8192 entries ($75\%$ compress ratio). Every arm trains for 100 steps and differs from the row above in the single field marked in bold.
}
\label{tab:method_ladder}
\small
\begin{tabular}{llllcc}
\toprule
& \multicolumn{3}{c}{Component} & \multicolumn{2}{c}{RULER 8K} \\
\cmidrule(lr){2-4} \cmidrule(lr){5-6}
Configuration & Always-attended & Loss & Head budget allocation & Avg. & NIAH \\
\midrule
Direct instantiation & none & LM & uniform & 4.33 & 2.00 \\
$+$ always-attended entries & \textbf{sinks $+$ window} & LM & uniform & 44.86\,\textcolor[HTML]{2A63D8}{\tiny$(+40.53)$} & 53.73 \\
$+$ distillation & sinks $+$ window & \textbf{KL} & uniform & 58.86\,\textcolor[HTML]{2A63D8}{\tiny$(+14.00)$} & 72.78 \\
$+$ head budget & sinks $+$ window & KL & \textbf{mass} & 67.60\,\textcolor[HTML]{2A63D8}{\tiny$(+8.74)$} & 86.17 \\
\midrule
\rowcolor{ExpReference}
Dense backbone (no eviction) & --- & --- & --- & 70.01 & 86.95 \\
\bottomrule
\end{tabular}
\end{table}

\begin{table}[t]
\centering
\caption{
    \textbf{Which choice, within each component.} Every arm is the reference recipe ($\dagger$) with a single field changed, and $\Delta$ is its Avg.\ against that reference. Two seeds of the reference itself differ by $0.28$, so a $\Delta$ of that size is not a difference. Setup and per-task scores are in Appendix~\ref{app:ablation}.
}
\label{tab:method_components}
\small
\begin{tabular}{llccr}
\toprule
Component & Choice & Avg. & NIAH & $\Delta$ \\
\midrule
Always-attended & none & 4.72 & 2.00 & \textcolor[HTML]{777777}{\tiny$-62.87$} \\
 & sinks only & 0.57 & 0.00 & \textcolor[HTML]{777777}{\tiny$-67.03$} \\
 & window only & 57.81 & 72.71 & \textcolor[HTML]{777777}{\tiny$-9.79$} \\
 & \cellcolor{ExpOurs}sinks $+$ window$^{\dagger}$ & 67.60 & 86.17 & --- \\
\midrule
Loss & LM loss (CE) & 51.50 & 64.65 & \textcolor[HTML]{777777}{\tiny$-16.09$} \\
 & LongCE & 53.01 & 61.99 & \textcolor[HTML]{777777}{\tiny$-14.59$} \\
 & forward KL & 66.84 & 83.88 & \textcolor[HTML]{777777}{\tiny$-0.76$} \\
 & reverse KL $+$ $0.1$ CE & 67.33 & 85.00 & \textcolor[HTML]{777777}{\tiny$-0.27$} \\
 & \cellcolor{ExpOurs}reverse KL$^{\dagger}$ & 67.60 & 86.17 & --- \\
\midrule
Scorer & $h_j$ only, linear ($0.5$M) & 53.53 & 63.21 & \textcolor[HTML]{777777}{\tiny$-14.07$} \\
 & \cellcolor{ExpOurs}$h_j$ only, MLP ($7.5$M)$^{\dagger}$ & 67.60 & 86.17 & --- \\
 & $+$ 8 preceding state ($16.8$M) & 66.82 & 85.67 & \textcolor[HTML]{777777}{\tiny$-0.78$} \\
 & $+$ all preceding recurrent state ($24.1$M) & 68.76 & 87.20 & \textcolor[HTML]{2A63D8}{\tiny$+1.16$} \\
 & $+$ all preceding attention ($19.46$M) & 66.63 & 85.48 & \textcolor[HTML]{777777}{\tiny$-0.97$} \\
 & fastkvzip-style ($11.36$M) & 67.25 & 84.76 & \textcolor[HTML]{777777}{\tiny$-0.35$} \\
\midrule
Head budget & uniform & 58.86 & 72.78 & \textcolor[HTML]{777777}{\tiny$-8.74$} \\
allocation & permuted & 57.14 & 68.92 & \textcolor[HTML]{777777}{\tiny$-10.46$} \\
 & offline table & 62.33 & 75.25 & \textcolor[HTML]{777777}{\tiny$-5.26$} \\
 & \cellcolor{ExpOurs}retained mass$^{\dagger}$ & 67.60 & 86.17 & --- \\
\midrule
Retention rate & off & 69.27 & 88.15 & \textcolor[HTML]{2A63D8}{\tiny$+1.67$} \\
 & \cellcolor{ExpOurs}on$^{\dagger}$ & 67.60 & 86.17 & --- \\
\bottomrule
\end{tabular}
\end{table}

\begin{table}[t]
\vspace{-3mm}
\centering
\caption{
    Train the backbone with gate decalibrates the gate rather than damage the model.
}
\label{tab:method_joint}
\small
\begin{tabular}{llrccc}
\toprule
& & & \multicolumn{2}{c}{RULER 8K} & \\
\cmidrule(lr){4-5}
Loss & Trained & Tokens & Evict 75\% & Full KV & Compression gap \\
\midrule
reverse KL & gate only$^{\dagger}$ & 7M & \textbf{67.60} & 70.01 & $-2.41$ \\
reverse KL & gate $+$ backbone & 7M & 62.83\,\textcolor[HTML]{777777}{\tiny$(-4.77)$} & 71.65 & $-8.82$ \\
LM loss & gate $+$ backbone & 7M & 60.45\,\textcolor[HTML]{777777}{\tiny$(-7.15)$} & 69.07 & $-8.62$ \\
LM loss & gate $+$ backbone & 79M & 59.68\,\textcolor[HTML]{777777}{\tiny$(-7.92)$} & 69.26 & $-9.57$ \\
LM loss & gate $+$ backbone & 197M & 57.42\,\textcolor[HTML]{777777}{\tiny$(-10.18)$} & 69.24 & $-11.82$ \\
\bottomrule
\end{tabular}
\vspace{-5mm}
\end{table}

\subsection{Learning a Reliable Eviction Scorer}
\paragraph{The most direct instantiation does not work.}
\label{sec:method:naive}
We first consider the most direct design of Equation~\ref{eq:paradigm}: an MLP maps each hidden state $h_j$ to a score $s_j$ and is trained with the language-modeling loss. This setup collapses to nearly constant scores, reaching $4.33$ where the dense model reaches $70.01$. If all $g_j$ take the same positive value, $\log g_j$ adds the same constant to every key logit, leaving the attention weights unchanged. The constant score is thus a no-op, and the objective prefers it: the frozen backbone was fit for dense attention, so its minimizer is a small perturbation of the constant score. Nearly constant scores are useless for eviction: the top-$B$ cut falls among near-ties, so the retained set is arbitrary (Appendix~\ref{app:methoddetails}).

\paragraph{Removing the no-op solution.}
\label{sec:method:pin}
\label{sec:method:budget}
The scores should not be able to reproduce dense attention. Eviction already keeps a few attention sinks and a causal local window under any policy, and these entries give the gate a reference frame: holding their gates at one fixes the scale, so the level of the remaining gates is no longer free. We then require the evictable entries to share a fixed total mass $\beta$, which they cannot do at the value that would recover dense attention. Writing $P_i$ for the protected set and $E_i$ for the evictable one,
\begin{equation}
g_{ij}=
\begin{cases}
1, & j\in P_i,\\[2pt]
\displaystyle \beta\frac{\exp(s_j)}{\sum_{\ell\in E_i}\exp(s_\ell)},
& j\in E_i,
\end{cases}
\qquad
a_{ij}\propto \exp\!\left(\frac{q_i k_j^\top}{\sqrt d}\right)g_{ij}.
\label{eq:pinned-gate}
\end{equation}
Dense attention is now out of the question: it needs $g_{ij}=1$ throughout $E_i$, hence $\beta=|E_i|$, whereas we set $\beta\ll|E_i|$. The scorer has to rank them (Appendix~\ref{app:methoddetails}). Because $\sum_{j\in E_i}g_{ij}=\beta$ holds exactly at every step, $\beta$ is also the sparsity used during training rather than a target approached through a penalty, which is what makes a sparsity regularizer or schedule unnecessary~\citep{trimkv, dms}. This mechanism raises pilot RULER from 4.33 to 44.86 (Table~\ref{tab:method_ladder}).

\paragraph{Distillation isolates the error caused by eviction.}
\label{sec:method:obj}
With the language-modeling loss, the design above reaches only $44.86$ because the cross-entropy gradient mixes errors caused by eviction with errors already made by the frozen backbone. Let $p_t^{\mathrm{gated}}$ and $p_t^{\mathrm{dense}}$ denote the gated and dense next-token probability distributions, and let $e_{y_t}$ be the one-hot target. The gradient with respect to the output logits decomposes as
\begin{equation}
p^{\text{gated}}_t-e_{y_t}
=\underbrace{\left(p^{\text{gated}}_t-p^{\text{dense}}_t\right)}_{\text{caused by eviction}}
+\underbrace{\left(p^{\text{dense}}_t-e_{y_t}\right)}_{\text{frozen-backbone error}} .
\label{eq:ce-decomposition}
\end{equation}
The gate can reduce only the first term, while the second is irreducible with a frozen backbone and can dominate at difficult, high-entropy positions. Token reweighting cannot remove this interference. LongCE~\citep{longce} usefully emphasizes tokens that benefit from long context, but still scales both terms in Equation~\ref{eq:ce-decomposition} together.

Distillation removes the second term by supervising the gate against the dense distribution. The gradient then depends only on $p^{\text{gated}}_t$ and $p^{\text{dense}}_t$, and vanishes once the two agree, so the frozen-backbone error never enters. Either direction of the KL has this property: the forward direction gives exactly the first term of Equation~\ref{eq:ce-decomposition}, while the reverse direction we adopt matches it to first order and differs only in how it treats the teacher's low-probability region. We choose between them empirically (Appendix~\ref{app:methoddetails}). Supervision also becomes dense, since each position contributes a full distribution over the vocabulary instead of a single target token. This is the $+14.00$ from $44.86$ to $58.86$ in Table~\ref{tab:method_ladder}.

\paragraph{Scores are not comparable across heads.}
\label{sec:method:headbudget}
Attention heads specialize in different local and long-range patterns, so they need different cache capacities, and a fixed offline budget table already improves over a uniform split, confirming that reallocating capacity is beneficial. But the retention score cannot decide that allocation. Adding a constant to every score of one head shifts its normalizer by the same amount, leaving both Equation~\ref{eq:pinned-gate} and the entries retained within that head unchanged, so the absolute level of a head's scores is not identifiable and scores from different heads are not comparable. The allocation must therefore be read off a quantity that this shift does not move, and the attention mass a head retains is such a quantity. It also depends on the current context, which the offline table cannot follow.
We therefore allocate the budget online from the final prefill query. For each head, we order the evictable entries by the retention score and compute the cumulative attention mass retained by its top $k$ entries, including the always-attended sinks and window. We then give every head the smallest $k_h$ that reaches a shared mass target $\rho^\star$, adjusting $\rho^\star$ and any integer remainder so that $\sum_h k_h$ exactly equals the uniform baseline's total budget. This assigns more slots to heads whose attention is distributed over more entries and raises the average from $58.86$ to $67.60$. Permuting the same set of per-head budgets reduces performance to $57.14$, below the uniform split, confirming that the gain comes from matching each budget to the head and context that need it, rather than from using unequal budgets alone (Table~\ref{tab:method_components}). This allocator can be viewed as budget-constrained per-head top-$p$, with the shared mass threshold chosen to satisfy the global cache budget.

\subsection{Scorer Architecture and Training Stability}
\begin{figure}[!htbp]
\centering
\includegraphics[width=\linewidth]{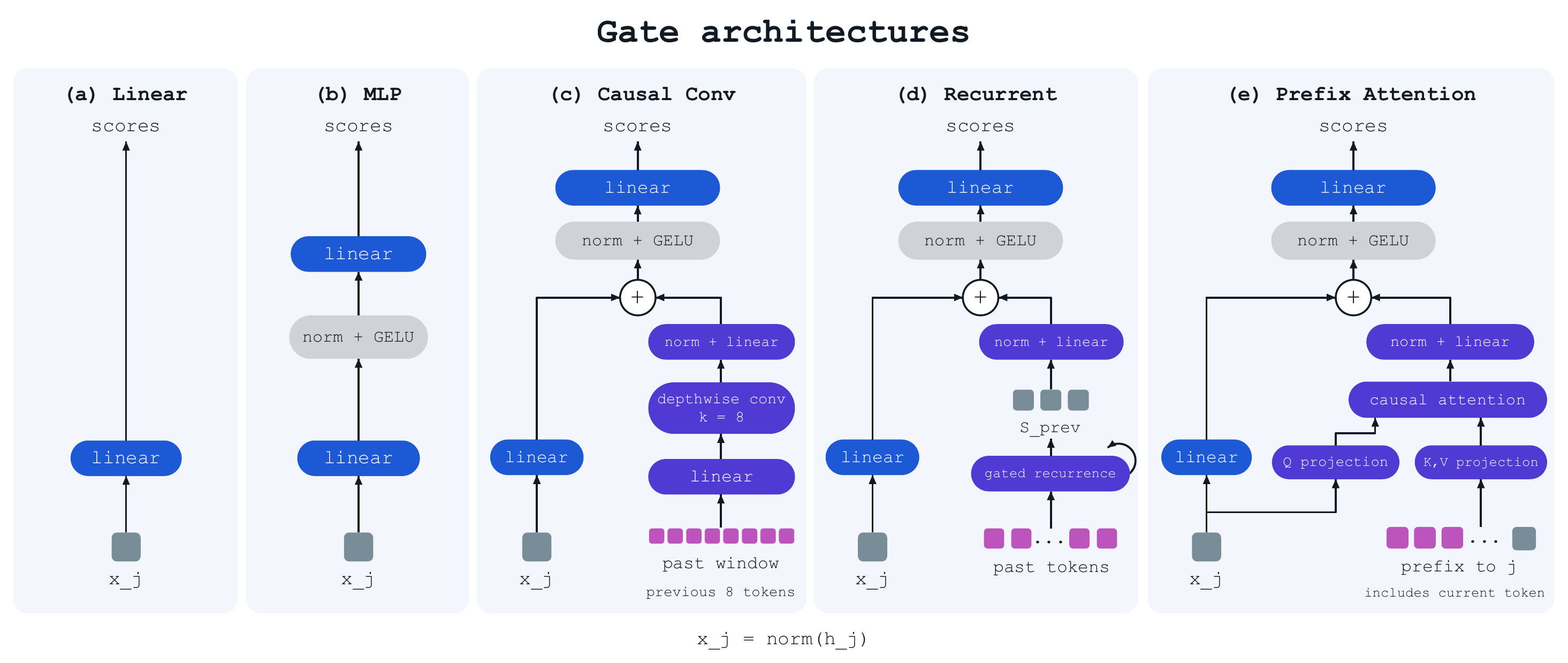}
\caption{\textbf{Scorer architectures.} All variants map $x_j=\operatorname{Norm}(h_j)$ to per-head retention scores and differ only in what they read besides $x_j$. (a) A linear readout and (b) the token-local MLP used as the reference scorer see the current hidden state alone. The three history variants add one branch that is summed into the MLP's pre-activation: (c) a depthwise causal convolution over the previous eight tokens, (d) a gated recurrence carrying a fixed-size state $S_{j-1}$, and (e) a single-head causal attention over the prefix up to $j$. The added branch is initialized to zero, so each variant starts exactly from (b).}
\label{fig:gate-architectures}
\end{figure}

\paragraph{The scorer needs capacity, but not a wider view of the past.}
\label{sec:method:scorer}
We next study the scorer architectures in Figure~\ref{fig:gate-architectures}. The scorer requires sufficient capacity: replacing the $7.5$M-parameter MLP with a $0.52$M linear readout costs $14.07$ points (Table~\ref{tab:method_components}). Because identifying redundancy may require comparing an entry with information already stored in the past, we augment the MLP with three forms of explicit history: an eight-token causal convolution, a recurrent state, and full-prefix attention. Despite increasing the scorer to as many as $24.1$M parameters, all three variants remain within $1.2$ points of the reference. This suggests that the contextualized hidden state $h_j$ already contains the historical information needed to estimate redundancy. A Fast-KVzip-style adaptive reference head~\citep{fastkvzip} also provides no benefit. We therefore use a two-layer MLP applied directly to $h_j$.

\paragraph{A per-key retention rate stabilizes the retention scores.}
\label{sec:method:decay}
A score that depends only on content has nothing anchoring its overall level, so scores can drift upward together as training proceeds and the choice of retained entries becomes sensitive to small differences among large values. Following TrimKV~\citep{trimkv}, we give each entry a learned lifetime alongside its magnitude,
\begin{equation}
s_j(i)=m_j+\log\beta_j\,\frac{i-j}{\tau},\qquad \log\beta_j\le 0,
\label{eq:retention}
\end{equation}
Query independence here refers to independence from future query content; the score may still vary with token age. The policy remains top-$B$ at every step, but because the ranking now drifts with $i$, an entry once evicted is never reconsidered even if a retained entry later decays below it (Appendix~\ref{app:biasfamily}). Disabling the retention rate reaches $69.27$ against $67.60$ (Table~\ref{tab:method_components}); we keep it for a stable, explicitly parameterized age scale rather than for a gain.

\paragraph{Training the backbone jointly decalibrates the scorer.}
\label{sec:method:joint}
SP-KV~\citep{spkv} jointly trains the language model and utility predictor, allowing sparsity to emerge during training. We do not observe the same benefit: unfreezing the attention projections and norms consistently reduces eviction performance. To distinguish backbone degradation from scorer failure, we also evaluate each checkpoint with Full-KV. Under reverse KL, Full-KV improves from $70.01$ to $71.65$ (Table~\ref{tab:method_joint}), while the eviction performance falls from $67.60$ to $62.83$. The backbone remains intact, but its changed representations decalibrate the scorer. We therefore keep the backbone frozen.

\subsection{Recovering Evicted Information}

\begin{wrapfigure}{r}{0.52\linewidth}
\vspace{-6pt}
\centering
\includegraphics[width=\linewidth]{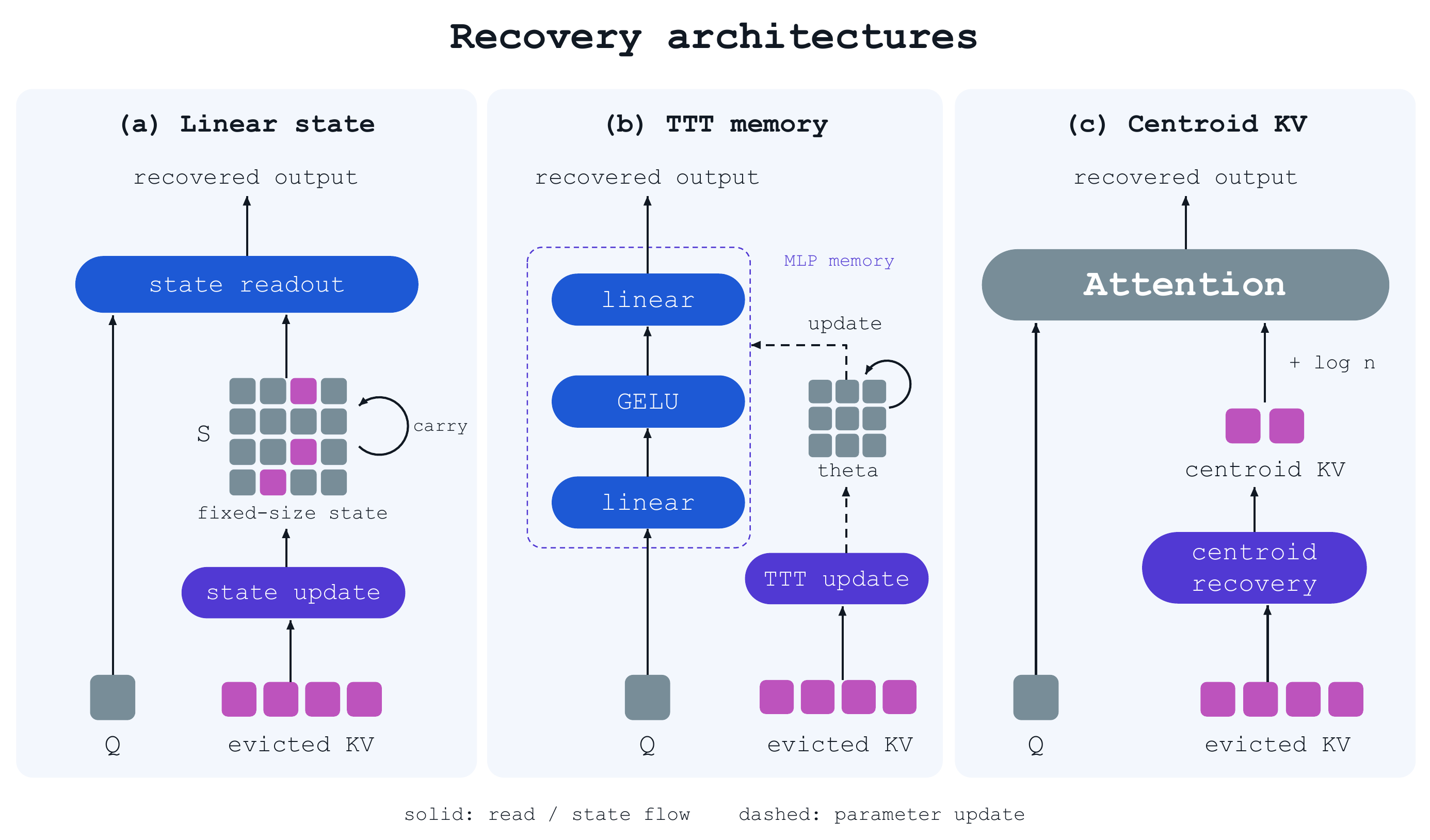}
\caption{\textbf{Recovery architectures.} Linear-state and TTT memories add recovery after attention; centroid KV competes with retained entries within the same softmax.}
\label{fig:recovery-architectures}
\vspace{-6pt}
\end{wrapfigure}

\paragraph{Compensating for what eviction discards.}
\label{sec:method:cmp}

Even accurate retention scores cannot make eviction lossless, since the retained entries may preserve part of an evicted entry's information but not all of it. With $o_S$ and $o_E$ the normalized outputs over the retained and evicted sets and $\rho=Z_E/(Z_S+Z_E)$ the mass on the evicted ones, dense attention gives $o=(1-\rho)o_S+\rho o_E$ while eviction returns $o_S$, an error of $\rho(o_E-o_S)$. It therefore grows with both how much mass is removed and how different the discarded content is from what remains, and we measure $\rho\approx0.2$, so the missing contribution is not negligible. Recovery needs both quantities, and they are not equally accessible: $\rho$ is a ratio of softmax normalizers, available where the gate was and only indirectly elsewhere. 
Since $o_E$ pools many entries, a fixed-size memory is the natural way to summarize it. We test a linear associative state $W:k\mapsto v$ and a TTT-style MLP fitted to the evicted pairs at inference, both read afterwards as $\alpha(q)o_S+\beta(q)m_E(q)$ (Figure~\ref{fig:recovery-architectures}). Measured by cosine similarity to the exact $o_E$, they reach only $-0.09$ to $0.44$, and enlarging them barely helps. The reason is not what they store but where they are read: a memory recovering $o_E$ exactly would still owe $\alpha=1-\rho$ and $\beta=\rho$, and a fusion placed after the softmax cannot obtain either from $Z_S$ and $Z_E$, so it has to learn the calibration indirectly in output space (Appendix~\ref{sec:recovering-evicted}).

\begin{wraptable}{r}{0.50\linewidth}
\vspace{-6pt}
\centering
\caption{\textbf{Centroid recovery at matched budgets.} The total cache budget $K$ includes recovery slots.}
\label{tab:method_cmp}
\begingroup
\fontsize{8}{9.5}\selectfont
\begin{tabular}{@{}llll@{}}
\toprule
Model \& Benchmark & $K$ & No slots & $64$ slots \\
\midrule
Qwen3-0.6B Math500 & 512 & 48.00 & \textbf{52.00}\,\textcolor[HTML]{2A63D8}{\tiny$(+4.00)$} \\
Qwen3-8B RULER 8K & 2048 & 76.65 & \textbf{77.15}\,\textcolor[HTML]{2A63D8}{\tiny$(+0.49)$} \\
\bottomrule
\end{tabular}
\par
\endgroup
\vspace{-6pt}
\end{wraptable}

The key is therefore to read the summary through attention's softmax. We cluster the evicted entries, average each cluster into a centroid key and value, and append the $R$ centroids to the retained cache, where $q_i k_c^\top/\sqrt d+\log n_r$ sets a centroid's weight and $v_c$ its output. The values supply $o_E$, the logits and cluster-size corrections supply $\rho$, and one softmax settles both. This parameter-free construction reaches cosine similarity of $0.85$--$0.96$ at most layers and improves Math500 and RULER 8K at matched budgets (Table~\ref{tab:method_cmp}). The Math500 gain is the larger one, consistent with compensation preserving information accumulated over a long CoT, whereas NIAH depends on retaining the correct evidence entry, which an average cannot substitute for.

\section{Experiments}
\label{sec:exp}
\subsection{Experimental Setup}
\label{sec:exp:setup}
\begingroup
\CompactParagraphs

\paragraph{Models and baselines.}
We evaluate Qwen3-4B and Qwen3-8B~\citep{qwen3} on RULER~\citep{hsieh2024ruler}, and Qwen3-8B on LongBench~\citep{bai2024longbench}, Math500~\citep{math500}, and AIME25, with uncompressed Full-KV as the reference throughout. The long-context baselines are SnapKV~\citep{li2024snapkv}, KeyDiff~\citep{keydiff}, DMA~\citep{shi2025dma}, and TrimKV~\citep{trimkv}, plus KVpop~\citep{hauzenberger2026kvpop} on RULER; the math baselines are R-KV~\citep{rkv}, TriAttention~\citep{triattention}, KVzap~\citep{kvzap}, Fast-KVzip~\citep{fastkvzip}, and Random Attention~\citep{randomattn}. DMA produces a query-independent score from value representations, which in our adaptation is added to the attention logits before the softmax during training and used for hard eviction at inference (Appendix~\ref{app:exp:protocol}). TrimKV uses its released gate weights. Unless otherwise specified, every method trained in this work updates only its gate parameters and keeps the backbone frozen.

\begin{table}[!t]
\centering
\vspace{-4mm}
\caption{\textbf{RULER overall results.}
Results for Qwen3-4B and Qwen3-8B with 8K and 16K contexts at compression ratios of 50\% and 75\%.
Full-KV uses the uncompressed cache, so its score is repeated in both compression columns.
The Qwen3-4B checkpoint of KVpop is not publicly available.}
\label{tab:exp-ruler-overview}

\begingroup
\fontsize{8.7}{10.6}\selectfont
\setlength{\tabcolsep}{3.0pt}
\renewcommand{\arraystretch}{1.16}
\setlength{\ExpMethodWidth}{0.205\linewidth}
\setlength{\ExpNumWidth}{
  \dimexpr(\linewidth-\ExpMethodWidth-16\tabcolsep)/8\relax
}

\resizebox{0.85\linewidth}{!}{%
\begin{tabular}{
  @{}p{\ExpMethodWidth}
  *{8}{>{\raggedleft\arraybackslash}p{\ExpNumWidth}}@{}
}
\toprule
& \multicolumn{2}{c}{\textbf{Qwen3-4B} (8K)}
& \multicolumn{2}{c}{\textbf{Qwen3-4B} (16K)}
& \multicolumn{2}{c}{\textbf{Qwen3-8B} (8K)}
& \multicolumn{2}{c}{\textbf{Qwen3-8B} (16K)} \\
\cmidrule(lr){2-3}\cmidrule(lr){4-5}
\cmidrule(lr){6-7}\cmidrule(lr){8-9}
Method & 50\% & 75\% & 50\% & 75\%
       & 50\% & 75\% & 50\% & 75\% \\
\midrule

\rowcolor{ExpReference}
Full-KV & 92.41 & 92.41 & 91.64 & 91.64
        & 93.86 & 93.86 & 92.96 & 92.96 \\
\midrule

SnapKV & 54.24 & 30.85 & 61.61 & 35.80
       & 54.32 & 35.13 & 62.74 & 41.54 \\
KeyDiff & 73.72 & 62.31 & 70.00 & 61.99
        & 77.35 & 67.14 & 74.44 & 66.87 \\
DMA & 68.18 & 36.34 & 64.69 & 33.30
    & 75.22 & 41.76 & 71.78 & 36.19 \\
TrimKV & 86.38 & 75.42 & 86.40 & 76.29
       & 87.23 & 80.26 & 85.91 & 79.09 \\
KVpop & \textemdash & \textemdash & \textemdash & \textemdash
      & 88.03 & 72.71 & 85.49 & 71.81 \\

\rowcolor{ExpOurs}
\textbf{IndexMem++}
& \textbf{91.75} & \textbf{83.89}
& \textbf{89.18} & \textbf{76.56}
& \textbf{93.79} & \textbf{90.14}
& \textbf{90.87} & \textbf{80.93} \\
\bottomrule
\end{tabular}%
}
\endgroup
\end{table}

\paragraph{Tasks.}
RULER (8K and 16K contexts) and LongBench use greedy decoding at 50\% and 75\% compression. For a given input and compression ratio, every method receives the same average per-KV-head budget, which is inclusive of attention sinks, the local window, and any auxiliary slots the method maintains; Appendix~\ref{app:exp:longcontext} gives the budget accounting and the protected region of each method. LongBench covers \ExpLongBenchTaskCount{} tasks with \ExpLongBenchSampleCount{} examples. We leave out TREC because its question-classification score is highly sensitive to prompt formatting and non-monotonic in the cache budget, which makes it a poor measure of compression quality. For math reasoning, we estimate pass@1 from two rollouts per Math500 problem and eight per AIME25 problem, with thinking enabled, temperature 0.6, and top-$p$ 0.95. Budgets, counted the same way, are 512 and 1{,}024 entries for Math500 and 1{,}024 and 2{,}048 for AIME25; Appendix~\ref{app:exp:math} lists the generation limits, seeds, and problem subsets of each run.

\paragraph{Training IndexMem++.}
We train only the gate, with the backbone frozen, on long documents from LongMino~\citep{longmino} under an 8K$\to$16K length curriculum. The gate is the configuration selected in Section~\ref{sec:method}, a two-layer MLP of width 256 on the hidden state with a per-key retention rate, trained by reverse-KL distillation against the same model with full KV; four attention sinks and a 128-token local window are always attended. The head-budget allocator (Section~\ref{sec:method:headbudget}) and the centroid slots (Section~\ref{sec:method:cmp}) act only at inference and add no trained parameters. On Math500 and AIME25 the prompt is short and the cache fills up during generation, so the allocator, which reads the attention at the end of prefill, has almost nothing to work from. There we use a head table fitted offline for each budget instead of the per-document allocator, and no centroid slots (Appendix~\ref{app:exp:protocol}).
\endgroup

\subsection{Long-Context Understanding}
\label{sec:exp:longcontext}
\label{sec:exp:ruler}
\begin{table}[!htbp]
\centering
\caption{\textbf{RULER task-level results at 8K and a compression ratio of 75\%.} Qwen3-4B and Qwen3-8B on all 13 RULER tasks.}
\label{tab:exp-ruler-tasks}
\begingroup\fontsize{7.8}{9.5}\selectfont
\setlength{\tabcolsep}{1.0pt}\renewcommand{\arraystretch}{1.13}
\resizebox{\linewidth}{!}{%
\begin{tabular}{@{}cl*{14}{r}@{}}
\toprule
\textbf{Model} & \textbf{Method} & \multicolumn{14}{c}{\textbf{RULER Tasks (8K)}} \\
\cmidrule(l){3-16}
 &  & S1 & S2 & S3 & MK1 & MK2 & MK3 & MV & MQ & VT & CWE & FWE & QA1 & QA2 & Avg. \\
\midrule
\rowcolor{ExpReference}
\cellcolor{white}\multirow{6}{*}{\rotatebox[origin=c]{90}{\textbf{Qwen3-4B}}} & Full-KV & 100.00 & 100.00 & 100.00 & 99.80 & 99.80 & 99.80 & 99.85 & 99.70 & 99.96 & 84.56 & 86.27 & 75.20 & 56.40 & 92.41 \\
\cmidrule(l){2-16}
 & SnapKV & 70.20 & 22.20 & 2.40 & 15.40 & 8.00 & 3.40 & 14.05 & 13.85 & 38.36 & 62.18 & \textbf{89.00} & 29.60 & 32.40 & 30.85 \\
 & KeyDiff & \textbf{100.00} & 88.60 & 92.00 & 74.60 & 4.60 & 0.00 & 89.90 & 89.50 & \textbf{100.00} & 69.14 & 74.07 & 15.00 & 12.60 & 62.31 \\
 & DMA & 99.00 & 24.20 & 2.40 & 27.40 & 5.40 & 0.00 & 14.30 & 20.70 & 79.60 & 48.16 & 71.47 & 47.80 & 32.00 & 36.34 \\
 & TrimKV & \textbf{100.00} & 98.80 & \textbf{92.60} & 95.00 & 48.40 & 10.80 & 91.55 & 95.60 & 99.96 & \textbf{77.00} & 71.60 & 49.60 & 49.60 & 75.42 \\
\rowcolor{ExpOurs}
\cellcolor{white} & \textbf{IndexMem++} & \textbf{100.00} & \textbf{99.20} & 91.20 & \textbf{97.80} & \textbf{94.80} & \textbf{60.40} & \textbf{98.40} & \textbf{99.20} & \textbf{100.00} & 44.28 & 84.07 & \textbf{68.00} & \textbf{53.20} & \textbf{83.89} \\
\midrule
\rowcolor{ExpReference}
\cellcolor{white}\multirow{7}{*}{\rotatebox[origin=c]{90}{\textbf{Qwen3-8B}}} & Full-KV & 100.00 & 100.00 & 100.00 & 99.80 & 99.60 & 99.60 & 100.00 & 100.00 & 100.00 & 94.24 & 91.00 & 75.80 & 60.20 & 93.86 \\
\cmidrule(l){2-16}
 & SnapKV & 76.00 & 21.20 & 2.40 & 16.00 & 5.80 & 4.40 & 10.70 & 12.40 & 66.00 & 78.32 & 89.33 & 36.20 & 38.00 & 35.13 \\
 & KeyDiff & \textbf{100.00} & 95.20 & 94.80 & 86.60 & 6.80 & 1.00 & 98.20 & 95.90 & \textbf{100.00} & 83.62 & 74.87 & 20.20 & 15.60 & 67.14 \\
 & DMA & 91.20 & 26.00 & 13.00 & 34.40 & 8.20 & 0.00 & 18.85 & 29.05 & 92.72 & 58.24 & 80.67 & 55.00 & 35.60 & 41.76 \\
 & TrimKV & \textbf{100.00} & \textbf{100.00} & \textbf{99.60} & 97.80 & 51.00 & 4.40 & 97.80 & 99.35 & \textbf{100.00} & 91.34 & 91.27 & 58.00 & 52.80 & 80.26 \\
 & KVpop & \textbf{100.00} & 94.20 & 83.20 & 79.20 & 37.80 & 2.00 & 89.55 & 89.70 & \textbf{100.00} & \textbf{92.06} & 86.93 & 44.20 & 46.40 & 72.71 \\
\rowcolor{ExpOurs}
\cellcolor{white} & \textbf{IndexMem++} & \textbf{100.00} & \textbf{100.00} & 96.40 & \textbf{99.20} & \textbf{97.80} & \textbf{73.00} & \textbf{98.80} & \textbf{100.00} & \textbf{100.00} & 83.18 & \textbf{91.40} & \textbf{71.40} & \textbf{60.60} & \textbf{90.14} \\
\bottomrule
\end{tabular}
}
\endgroup
\par\vspace{3pt}\begin{minipage}{\linewidth}\fontsize{8}{9.3}\selectfont S1--S3 = single-needle tasks; MK1--MK3 = multi-key tasks; MV = multi-value; MQ = multi-query; VT = variable tracking; CWE/FWE = common/frequent word extraction; QA1/QA2 = question answering.\end{minipage}
\end{table}

\paragraph{RULER highlights the importance of identifying redundancy.}
The prefill cache is compressed before the target question is available. SnapKV scores relevance from the attention of the queries it has seen, and recent-context attention says little about what a future query will need, which is why it degrades substantially on the retrieval tasks. Methods that score redundancy directly, including KeyDiff, TrimKV, and KVpop, do better. A single needle is easy to distinguish from filler text; MultiKey-2/3 are not, since at 8K they hold 425 and 125 distractor records sharing the target's template and key--value type, and with the question still unseen every record is potentially relevant and no distractor can be ruled out in advance. This is where similarity-based criteria degrade sharply and IndexMem++ remains robust.
DMA also degrades sharply at high compression. Its score is query-independent, as ours is, but it is computed from the values and trained with the language-modeling loss rather than distilled from the Full-KV model, and the resulting ranking transfers poorly to hard eviction on the retrieval-heavy tasks.

\label{sec:exp:longbench}
\begin{table}[!htbp]
\centering
\caption{\textbf{LongBench results on Qwen3-8B.} Category means and the average over 15 tasks, at 50\% and 75\% compression.}
\label{tab:exp-longbench}
\begingroup\fontsize{8.5}{10.3}\selectfont
\setlength{\tabcolsep}{4pt}\renewcommand{\arraystretch}{1.15}
\resizebox{0.85\linewidth}{!}{%
\begin{tabular}{@{}lc*{7}{r}@{}}
\toprule
\textbf{Method} & \textbf{Compression ratio} & \textbf{Single QA} & \textbf{Multi QA} & \textbf{Summ.} & \textbf{Few-shot} & \textbf{Synth.} & \textbf{Code} & \textbf{Avg.} \\
\midrule
\rowcolor{ExpReference}
Full-KV &  & 41.96 & 48.58 & 27.61 & 65.40 & 50.30 & 62.47 & 47.39 \\
\midrule
SnapKV & 50\% & 35.60 & 46.52 & 25.93 & 65.26 & 50.00 & \textbf{63.10} & 45.39 \\
KeyDiff & 50\% & 28.87 & 28.91 & 24.46 & 62.27 & 45.61 & 45.61 & 36.91 \\
DMA & 50\% & 38.87 & 42.72 & 26.53 & 64.33 & 35.25 & 55.13 & 42.25 \\
TrimKV & 50\% & \textbf{41.93} & 46.29 & 27.20 & \textbf{67.81} & 51.75 & 56.35 & 46.54 \\
\rowcolor{ExpOurs}
\textbf{IndexMem++} & 50\% & 40.82 & \textbf{49.06} & \textbf{27.64} & 65.27 & \textbf{52.61} & 62.32 & \textbf{47.53} \\
\midrule
SnapKV & 75\% & 28.59 & 39.69 & 24.12 & 65.04 & 47.92 & \textbf{63.62} & 42.02 \\
KeyDiff & 75\% & 20.14 & 19.35 & 19.12 & 58.91 & 26.58 & 38.03 & 28.19 \\
DMA & 75\% & 29.17 & 34.39 & 23.54 & 59.69 & 5.00 & 50.09 & 32.72 \\
TrimKV & 75\% & \textbf{39.64} & 44.58 & \textbf{27.00} & \textbf{66.45} & \textbf{51.50} & 55.84 & 45.41 \\
\rowcolor{ExpOurs}
\textbf{IndexMem++} & 75\% & 38.69 & \textbf{47.43} & 26.22 & 64.47 & 50.96 & 61.26 & \textbf{46.03} \\
\bottomrule\end{tabular}%
}
\endgroup
\par\vspace{3pt}\begin{minipage}{\linewidth}\fontsize{8}{9.3}\selectfont QA = question answering; Summ. = summarization; Synth. = synthetic tasks. Few-shot is the mean of TriviaQA and SAMSum; Avg. is the equal-weight mean of all 15 tasks.\end{minipage}
\vspace{-5mm}
\end{table}

\paragraph{LongBench highlights the importance of information recoverability.}
IndexMem++ achieves the highest overall average at both compression ratios and consistently leads on multi-document QA. The document-QA results separate information recoverability from proxy importance: low recent attention in SnapKV does not mean an entry has become redundant, the geometric distinctiveness KeyDiff relies on says nothing about whether the same information is represented elsewhere in the cache, and TrimKV's bias toward recent tokens ignores that the evidence a later question needs may sit anywhere in the context. These mismatches matter most in multi-document QA, where a correct answer may depend on several pieces of evidence spread across documents and positions, none of which need be recent or geometrically distinctive. At 75\% compression, IndexMem++ scores 61.57, 44.82, and 35.89 on HotpotQA, 2WikiMQA, and MuSiQue, against 59.53, 42.40, and 31.81 for TrimKV, with KeyDiff further behind, and it leads this category at both compression ratios. This is consistent with IndexMem++ better preserving evidence that remains irreplaceable despite being old, recently unattended, or geometrically unremarkable.

The two synthetic tasks stress complementary forms of information preservation. PassageRetrieval requires enough semantic detail to tell the target passage from competing ones, whereas PassageCount requires global coverage of the unique passages despite repeated content. DMA degrades on both under hard eviction, as on RULER, most visibly on PassageRetrieval at 75\% compression, where it drops to 8.50 while IndexMem++ retains 94.42; on PassageCount the corresponding scores are 1.50 and 7.50. IndexMem++ trains its scorer against the full-KV prediction distribution, which gives entries whose attenuation changes the model's predictions a direct signal to raise their retention priority. Appendix~\ref{app:exp:longbench} reports the complete per-task results.

\subsection{Preserving Reasoning State under KV-Cache Compression}

\begin{table}[!htbp]
\centering
\caption{\textbf{Mathematical reasoning with Qwen3-8B.} Each cell reports pass@1, with the mean number of generated tokens shown in smaller gray text.}
\label{tab:exp-math}
\begingroup
\fontsize{8.8}{10.6}\selectfont
\setlength{\tabcolsep}{5.0pt}
\renewcommand{\arraystretch}{1.16}
\newcommand{\MathResult}[2]{#1\,{\fontsize{6.6}{7.4}\selectfont\color{gray}(#2)}}
\begin{tabular}{@{}lrrrr@{}}
\toprule
& \multicolumn{2}{c}{\textbf{Math500}} & \multicolumn{2}{c}{\textbf{AIME25}} \\
\cmidrule(lr){2-3}\cmidrule(l){4-5}
\textbf{Method} & $K{=}1024$ & $K{=}512$ & $K{=}2048$ & $K{=}1024$ \\
\midrule
\rowcolor{ExpReference}
Full-KV & \multicolumn{2}{c}{\MathResult{0.832}{5,258}} & \multicolumn{2}{c}{\MathResult{0.700}{17,845}} \\
\midrule
KVzap & \MathResult{0.792$^\dagger$}{6,326} & \MathResult{0.548$^\dagger$}{9,850} & \MathResult{0.446}{22,019} & \MathResult{0.183}{29,013} \\
Fast-KVzip & \MathResult{0.736$^\dagger$}{6,974} & \MathResult{0.480$^\dagger$}{11,146} & \MathResult{0.375}{23,904} & \MathResult{0.154}{29,462} \\
R-KV & \MathResult{0.811}{9,741} & \MathResult{0.605}{12,757} & \MathResult{0.229}{27,044} & \MathResult{0.121}{29,625} \\
TriAttention & \MathResult{0.863}{8,563} & \MathResult{0.758}{12,056} & \MathResult{0.383}{23,109} & \MathResult{0.192}{28,507} \\
Random Attention & \MathResult{\textbf{0.869}}{7,342} & \MathResult{0.784}{9,714} & \MathResult{0.388}{23,199} & \MathResult{0.221}{27,538} \\
\rowcolor{ExpOurs}
\textbf{IndexMem++} & \MathResult{0.824}{5,226} & \MathResult{\textbf{0.794}}{5,423} & \MathResult{\textbf{0.600}}{18,180} & \MathResult{\textbf{0.450}}{20,279} \\
\bottomrule
\end{tabular}
\endgroup
\par\vspace{3pt}
\begin{minipage}{\linewidth}
\fontsize{8}{9.3}\selectfont
Math500 uses two rollouts per problem, and AIME25 I+II uses eight. $^\dagger$Evaluated on a 125-problem subset of Math500 (Appendix~\ref{app:exp:math}).
\end{minipage}
\end{table}

The strong performance of Random Attention suggests that, once the problem prompt is protected, many intermediate chain-of-thought states are redundant. At the lower compression setting on Math500 ($K=1024$), Random Attention is more accurate than IndexMem++ (0.869 vs.\ 0.824) but needs substantially longer generation (7,342 vs.\ 5,226 tokens), and at $K=512$ the order flips, with IndexMem++ both more accurate (0.794 vs.\ 0.784) and more efficient (5,423 vs.\ 9,714 tokens). On AIME25 the gap is wide, 0.600 vs.\ 0.388 at $K=2048$ and 0.450 vs.\ 0.221 at $K=1024$. IndexMem++ also stays close to Full-KV in generation length, 5,226 and 5,423 tokens on Math500 against 5,258, and 18,180 and 20,279 tokens on AIME25 against 17,845, whereas every other method generates longer outputs at the same budget. This suggests that IndexMem++ preserves the reasoning states needed under high compression rather than forcing the model to compensate with longer chains of thought.

\subsection{What IndexMem++ Retains}
\label{sec:exp:retention}
We examine whether IndexMem++ retains early context or reduces to a sliding window. The control is a recency policy that keeps the most recent entries under the same per-head budgets and centroid slots. Both run with physical eviction on Qwen3-8B at $K=2{,}048$, on long AIME25 reasoning traces and on software-engineering agent trajectories (SWE); Appendix~\ref{app:exp:retention} gives the setup.

IndexMem++ does not reduce to a sliding window. When eight traces per domain are fed token by token up to 8K, only 52.5\% (AIME25) and 48.4\% (SWE) of the entries it retains outside the sinks and local window fall inside a recency window of the same size, against 100\% for the recency policy. Early facts also stay cached until they are needed. In 17 of the 25 SWE trajectories, a fact from the issue description is applied later on, 39 facts in all; every one of them is still cached at the point of use, whereas the recency policy has already dropped the original entries of 37. Under the same budget, IndexMem++ also predicts the 256 assistant tokens after the 8K prefix better than the recency policy on every trace, with a lower negative log-likelihood (NLL) of 0.207 vs.\ 0.279 on AIME25 and 1.221 vs.\ 1.551 on SWE. In free generation on AIME25, it reaches a pass@1 of 0.592, against 0.121 for the recency policy and 0.654 for Full-KV.

\ifPDFTeX\let\scshape\upshape\fi
\subsection{Decoding Efficiency under a Fixed Budget}
\label{sec:exp:efficiency}
\begin{figure}[htbp]
  \centering
  \includegraphics[width=0.90\linewidth]{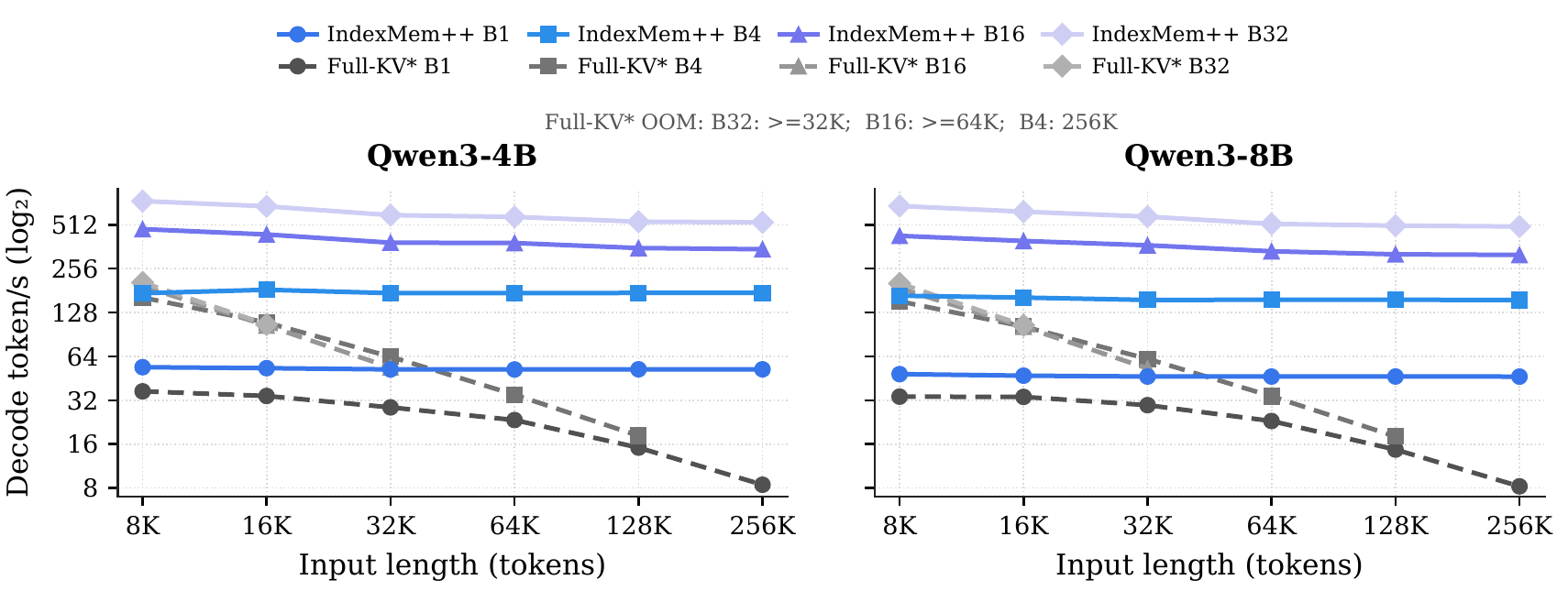}
  \vspace{-2mm}
  \caption{Decode throughput versus input length for Qwen3-4B and Qwen3-8B. Full-KV* is the Full-KV reference measured under the protocol of Appendix~\ref{app:efficiency}.}
  \label{fig:exp-decode-throughput}
  \vspace{-2mm}
\end{figure}
Figure~\ref{fig:exp-decode-throughput} reports batch throughput for 256 decode steps on a single H20 GPU at an average per-KV-head budget of 2,048 entries, excluding prefill and setup costs; the missing Full-KV* points are out-of-memory runs, and Appendix~\ref{app:efficiency} gives the full protocol. \textbf{IndexMem++ largely decouples decoding throughput from the input sequence length.} Continuous eviction bounds the historical KV cache of each sequence by a fixed budget, so the per-step KV reads and attention computation depend on that budget rather than on the sequence length, whereas Full-KV must read a cache that keeps growing. At a batch size of 32, increasing the input length from 8K to 256K ($32\times$) reduces the throughput of IndexMem++ by only about 28\% on both models.

\section{Related work}
\label{sec:related}


\paragraph{Learnable eviction gates.}
Learnable eviction replaces fixed heuristics with trainable, content-dependent scores. TrimKV~\citep{trimkv} predicts per-key retention scores that decay with age, while DMC~\citep{dmc} and DMS~\citep{dms} learn discrete token-level compression decisions through continuous relaxations and explicit budget objectives. SP-KV~\citep{spkv} predicts future utility to gate long-range attention, whereas Locret~\citep{locret} and Attention-Gate~\citep{attngate} train auxiliary modules to score entries during prefill.
Although motivated differently, these methods share the same paradigm: a lightweight scorer assigns retention scores to KV entries. We study the scorer, gate normalization and budget, supervision, and allocation across heads in Sections~\ref{sec:prelim} and~\ref{sec:method}.

\paragraph{Recovering evicted information.}
Summarizing a discarded past is a long-standing idea: compressive memories retain a coarse record of activations that leave the attention window~\citep{compressive}, gist tokens compress a prompt into a few learnable slots~\citep{gist}, and recurrent or test-time-trained states have been used as memory alongside attention~\citep{ttt,titans}. Such a summary can be read back either as a residual added to the attention output, whose scale is set by a gain learned outside the softmax, or as additional participants in attention, which compete for mass in the same softmax and are thus normalized by construction. We find this readout choice matters more than the capacity of the summary (Section~\ref{sec:method}).

\section{Conclusion}
Asked which of its own KV entries are expendable, a gate trained on the model's own loss answers that keeping everything is preferable, and returns constant scores. The question becomes answerable once that answer is unavailable: the evictable entries compete for a fixed gate mass, and the gate matches the dense model rather than the next token. What the score cannot say, how much budget each head deserves, is read off the mass each head retains. Some information is lost even so, and there the lesson is not to build a larger memory but to read the small one through the same softmax as the cache it supplements.

\section{Future work}
Our gate is trained on dense or teacher-forced prefixes but applied to a cache its own earlier decisions have compressed. On-policy distillation, with prefixes generated under hard eviction, would expose it to the state distribution it actually meets at inference.

\subsection*{AI use statement}

We used generative AI tools in two roles. First, as coding assistants during the development of our training and evaluation code, including the fused training-time attention kernel of Appendix~\ref{app:gatedattn} and the paged decode path of Appendix~\ref{app:hardevict}. All such code was reviewed by the authors, and every reported number comes from runs of code we have inspected; correctness of the kernels was additionally checked against reference implementations, as described in the corresponding appendices. Second, as a writing aid for improving the clarity and grammar of the manuscript. We did not use generative AI tools to generate research ideas, to design the method, to produce or modify experimental results, or to write the analysis of those results. We take responsibility for the final content of this work, including all text, claims, and artifacts.

\subsection*{Ethics statement}

This work studies KV-cache eviction for long-context inference. Our experiments use publicly released model checkpoints and public benchmarks, and involve no human subjects, no personally identifying information, and no newly collected data. We release no new dataset or model. The method reduces the memory and bandwidth cost of serving long-context models, which lowers the cost of deployment; it does not change what those models can be asked to do, and so inherits the capabilities and the risks of the underlying checkpoints rather than introducing new ones. One consideration specific to eviction deserves mention: because compression discards part of the cache, a compressed model can lose evidence that the uncompressed model would have used, and the resulting errors are not always visible to the user. Deployments in settings where faithfulness to the provided context matters should therefore validate accuracy at the compression ratio they intend to use, rather than assuming the ratios reported here transfer.

\subsection*{Reproducibility statement}

Section~\ref{sec:exp:setup} states the models, baselines, benchmarks, and compression ratios; Appendix~\ref{app:exp:protocol} gives the complete training recipe, including the training corpus and its subsets, the length curriculum, the optimizer schedule, the gate configuration, the distillation objective, and the inference-time components that add no trained parameters. Appendix~\ref{app:exp:longcontext} reports per-task results for all RULER and LongBench tasks, together with the budget accounting, retention criterion, and protected region used for each method, so that the budget each baseline receives is explicit. Appendix~\ref{app:ablation} gives the protocol for the pilot ablations of Section~\ref{sec:method}, including the seeds, and states which tasks carry no usable signal at that model scale. Appendices~\ref{app:gatedattn} and~\ref{app:hardevict} describe the training kernel and the conversion from soft gating to hard eviction, with the core routines listed as code. Appendix~\ref{app:efficiency} gives the measurement protocol for the decoding-throughput comparison, including the decode implementation used for each curve.



\bibliography{iclr2027_conference}
\bibliographystyle{plainnat}

\newpage
\appendix
\addtocontents{toc}{\protect\setcounter{tocdepth}{2}}
\renewcommand{\contentsname}{Appendix contents}
\tableofcontents
\newpage
\section{Other related work}
\label{app:related}

This appendix discusses other works that Section~\ref{sec:related} did not claim in detail.

\subsection{Training-free eviction.}
Training-free policies differ mainly in the signal they use to score entries. The largest group derives importance from attention already observed. H$_2$O retains heavy hitters by accumulating attention over past queries~\citep{zhang2023h2o,d2o}, SnapKV pools attention within a trailing observation window to decide what a prompt will need~\citep{li2024snapkv}, and R-KV adds a redundancy term so that repetitive reasoning traces are not retained many times over~\citep{rkv}. Because these scores depend on which queries have been seen, the induced ranking is revised as decoding proceeds, while the deletion it triggers is permanent.

A second group avoids queries altogether and scores entries from the statistics of the keys themselves. TriAttention compares keys against calibrated pre-RoPE query centers in the frequency domain~\citep{triattention}, while the $L_2$ key norm~\citep{keynorm} and KeyDiff's angular diversity criterion~\citep{keydiff} use the magnitude and the mutual similarity of the key vectors. These criteria are query-independent, which suits the irreversibility of eviction, but the statistic itself is fixed by hand. A third group uses position alone: StreamingLLM retains attention sinks together with a sliding window~\citep{xiao2024streamingllm}, and uniformly random retention serves as the natural control for how much of the benefit is attributable to the scoring at all~\citep{randomattn}. None of these methods requires training, which is their principal appeal; the cost is that a hand-chosen signal cannot adapt to what a particular context happens to make redundant.

\subsection{Sparse attention.}
Sparse attention also computes attention over a subset of entries, but it retains the full cache and chooses a new subset for every query, so no memory is reclaimed and no decision is irreversible. Training-free variants select blocks by cheap surrogates of the attention score~\citep{quest,minference}, and a growing body of work trains the selector jointly with the model~\citep{nsa,moba,dsa,shi2025dma,hils}. The trained selectors are close to our setting in machinery: they too must pass gradients through a discrete choice, and some study explicitly how a block-level gate should be injected into the attention computation and normalized~\citep{sas}. They are far from our setting in objective, however, because a selection that a query gets wrong there can be revised at the next query, whereas an eviction cannot.

\subsection{Gating in attention.}
\label{sec:gating}


The term \emph{gate} is used for several mechanisms at different stages of attention, which have different physical meanings. Let
\begin{equation}
    s_{ij} = \frac{q_i^\top k_j}{\sqrt d}, \qquad
    a_{ij} = \frac{\exp(s_{ij})}{\sum_{\ell}\exp(s_{i\ell})}, \qquad
    o_i = \sum_j a_{ij} v_j .
\end{equation}
A gate may act on the query, key, value, normalized attention output, or directly on the logits:
\begin{align*}
    \text{query gate:}\quad
        &q_i' = g_i^{Q}\odot q_i,
        &&\text{controls how the current token retrieves;} \\
    \text{key gate:}\quad
        &k_j' = g_j^{K}\odot k_j,
        &&\text{controls how a memory is addressed;} \\
    \text{value gate:}\quad
        &v_j' = g_j^{V}\odot v_j,
        &&\text{controls how much payload a memory returns;} \\
    \text{output gate:}\quad
        &o_i' = g_i^{O}\odot \sum_j a_{ij}v_j,
        &&\text{controls whether the retrieved message is used;} \\
    \text{logit gate:}\quad
        &a_{ij}' =
        \softmax_j\!\left(s_{ij} + b_{ij}\right),
        &&\text{controls whether query $i$ may read key $j$;} \\
    \text{temperature gate:}\quad
        &a_{ij}' =
        \softmax_j\!\left(\gamma_j s_{ij}\right),
        &&\text{controls how sharply key $j$ competes;} \\
    \text{denominator gate:}\quad
        &a_{ij}' = \frac{\exp(s_{ij})}{Z_i+c_i},
        &&\text{controls how much mass leaves the row.}
\end{align*}

These locations are not interchangeable. In particular, a gate outside the softmax changes the magnitude of an already-normalized message but does not reallocate probability mass. For example, value gating gives
\begin{equation*}
    o_i^{V}
    =
    \sum_j
    \underbrace{
    \frac{\exp(s_{ij})}{\sum_\ell \exp(s_{i\ell})}
    }_{a_{ij}}
    g_j^V v_j .
\end{equation*}
Even if $g_j^V\rightarrow 0$, key $j$ may still consume substantial attention probability; its mass is simply discarded rather than reassigned. The same distinction applies to the gated-attention mechanism of \citep{gatedattn}, which multiplies the output of scaled dot-product attention by a learned sigmoid,
\begin{equation*}
    o_i^{\mathrm{gate}}
    =
    g_i^O \odot
    \left(
    \sum_j a_{ij}v_j
    \right).
\end{equation*}
Besides providing a non-linearity between $W_V$ and $W_O$, this gate has been reported to suppress massive activations and attention sinks. Its role is nevertheless an \emph{output gate}: it can decide that the result of a retrieval should contribute little, but it does not decide which memory receives the released attention mass.
Related analyses of attention and residual sinks show that a small number of positions can absorb a disproportionate fraction of the normalized mass~\citep{sinks}; this is also why entries that must remain permanently accessible require explicit treatment in an eviction mechanism.

By contrast, a gate \emph{inside} the softmax acts directly on competition between memories.  Writing
\begin{equation}
    b_{ij} = \log g_{ij}, \quad g_{ij} > 0, \quad
    a_{ij}'= \frac{\exp(s_{ij}+b_{ij})}
         {\sum_\ell \exp(s_{i\ell}+b_{i\ell})}
    \nonumber
    =
    \frac{
        \exp(s_{ij})\, g_{ij}
    }{
        \sum_\ell \exp(s_{i\ell})\,g_{i\ell}
    } .
    \label{eq:gate-in-softmax}
\end{equation}
Thus an additive bias in logit space is exactly a multiplicative gate on the \emph{unnormalized} attention weights, followed by renormalization. Suppressing one entry therefore transfers its probability mass to the remaining entries.  In the hard limit $g_{ij}\rightarrow 0$, $b_{ij}\rightarrow-\infty$, recovering an attention mask. This makes the in-softmax location qualitatively different from query, key, value, or output gating: it is the natural location for a \emph{read gate} and hence for eviction.

Two other mechanisms also act inside the softmax but are unsuitable as retention scores. A multiplicative gate on the logit, $s_{ij}\leftarrow\gamma_js_{ij}$, is a per-key temperature, and a scalar key gate is the same operation because $g_j^Kq_i^\top k_j$ rescales the logit rather than shifting it. Shrinking $\gamma_j$ pulls the logit towards zero, which attenuates the entry where $s_{ij}>0$ but amplifies it where $s_{ij}<0$. Its effect therefore depends on the sign of the logit under the current query, so a single $\gamma_j$ does not order entries by how much they should be retained, whereas $b_j$ shifts every logit of key $j$ in the same direction regardless of the query. Key-Value Means uses such temperatures, but to control the sharpness of already-chosen regions rather than to select among them~\citep{kvm}.

A second mechanism modifies only the normalizer, $a_{ij}=\exp(s_{ij})/(Z_i+c_i)$, as in sigmoid attention, a null or register slot, and the sink construction analyzed by~\citep{sinks}. The mass $c_i/(Z_i+c_i)$ leaves the row without being assigned to any cached entry, so unlike an additive gate this suppresses an entry without making any other entry more visible. This is useful for absorbing mass that no entry should receive, which is why sinks need explicit treatment, but it provides no per-key quantity to rank.

\begin{table}[t]
\centering
\small
\setlength{\tabcolsep}{5pt}
\begin{tabular}{lccc}
\toprule
\textbf{Gate location}
& \textbf{Form}
& \textbf{Physical meaning}
& \textbf{Redistributes mass?} \\
\midrule
Query
& $q_i\!\leftarrow g_i^Q\odot q_i$
& retrieval mode
& Yes \\
Key
& $k_j\!\leftarrow g_j^K\odot k_j$
& address / retrievability
& Yes, non-monotonically \\
Value
& $v_j\!\leftarrow g_j^V\odot v_j$
& returned content
& No \\
SDPA output
& $o_i\!\leftarrow g_i^O\odot o_i$
& whether retrieval is used
& No \\
Logit, multiplicative
& $s_{ij}\!\leftarrow \gamma_j s_{ij}$
& competition sharpness
& Non-monotonically \\
Denominator only
& $Z_i\!\leftarrow Z_i+c_i$
& mass leaves the row
& No \\
Logit, additive
& $s_{ij}\!\leftarrow s_{ij}+b_{ij}$
& read visibility
& \textbf{Yes} \\
\bottomrule
\end{tabular}
\caption{
Gates at different locations in attention have different semantics. Among per-entry retention mechanisms, an additive logit gate provides monotonic control over normalized attention mass: lowering one entry's bias lowers its weight and transfers mass to the others. A scalar key gate is the multiplicative logit gate, since $g_j^Kq_i^\top k_j$ rescales the logit, and its effect on an entry's weight depends on the sign of that logit. A denominator gate removes mass from the row without assigning it to any entry.
}
\label{tab:gate-locations}
\end{table}

\paragraph{Additive in-softmax biases beyond eviction.}
\label{app:biasfamily}

An additive bias inside the softmax is not specific to eviction. The same mechanism is used to encode a broad range of architectural inductive biases:
\begin{equation}
    a_{ij}
    =
    \softmax_j\!\left(
        s_{ij} + b_{ij}
    \right).
    \label{eq:bias-family}
\end{equation}
Examples include fixed positional slopes~\citep{alibi}, learned token lifetimes~\citep{expirespan}, cumulative forget gates~\citep{fox}, pairwise cumulative suppression over intervening tokens~\citep{selectiveattn,dcp}, and query-conditioned allocations of attention mass~\citep{stickbreaking}. These mechanisms differ primarily in what information $b_{ij}$ is allowed to depend on.

Softmax is invariant to a row-wise offset:
\begin{equation}
    \softmax_j(s_{ij}+b_{ij})
    =
    \softmax_j(s_{ij}+b_{ij}+c_i).
\end{equation}
Hence $b_{ij}$ is only defined up to a query-dependent constant. Removing this irrelevant degree of freedom gives a useful \emph{canonical form}. For eviction, the key distinction is whether that canonical form still depends on the query index $i$. If not, the gate assigns every cached entry a persistent scalar ordering; if it does, the ordering may evolve over time and an entry that is weak now can become competitive again later.

We summarize the relevant bias families in Table~\ref{tab:biasfamily}. We additionally define the \emph{fold width} as the smallest $r$ such that, up to a row constant,
\begin{equation}
    b_{ij} = u_i^\top v_j,
    \qquad
    u_i,v_j\in\mathbb{R}^r .
\end{equation}
A small fold width allows the bias to be evaluated from compact per-query and per-key state without materializing an $n\times n$ matrix.

\begin{table}[t]
\centering
\small
\setlength{\tabcolsep}{4pt}
\begin{tabular}{lcccc}
\toprule
\textbf{Bias family}
& \textbf{$b_{ij}$}
& \textbf{Canonical form}
& \textbf{Frozen gate ranking?}
& \textbf{Fold width} \\
\midrule
Per-key scalar
& $\phi(j)$
& $\phi(j)$
& Yes
& $1$ \\

Separable lifetime
& $c_i-c_j$
& $-c_j$
& Yes
& $1$ \\

Key-specific decay
& $s_j+\frac{i-j}{\mathrm{ref}}\log\beta_j$
& depends on $i$
& No
& $2$ \\

Pairwise cumulative
& $\sum_{j<k<i} S_{kj}$
& depends on $i$
& No
& high \\

Query-conditioned
& $f(q_i,k_j)$
& depends on $i$
& No
& -- \\
\bottomrule
\end{tabular}
\caption{
Additive in-softmax biases viewed through their canonical form.
The second row includes forms such as the linear part of ALiBi and
cumulative forgetting that can be written as a difference of prefix
potentials; the third captures token-specific lifetime decay as used by
methods such as TrimKV and by the per-key retention rate of
Equation~\ref{eq:retention}; the fourth and fifth capture richer pairwise
or query-conditioned biases.  Only biases whose canonical form is key-only induce a persistent ordering; the remaining rows can change the ordering of memories over time.
}
\label{tab:biasfamily}
\end{table}

This distinction has three consequences for eviction. Our own gate belongs
to the third row of Table~\ref{tab:biasfamily} rather than the first: the
magnitude term $m_j$ alone would be a per-key scalar, but the retention
rate of Equation~\ref{eq:retention} makes the canonical form depend on the
query index. We state each consequence in general and then say where it
places our gate.

\paragraph{Persistent ordering.}
A canonical bias that does not depend on $i$ provides a persistent
ordering of cached entries.  Once an entry falls below a fixed keep budget
according to this gate, advancing the sequence cannot by itself make its
gate score overtake another already-existing entry.  Under this condition
the greedy online policy of Section~\ref{sec:prelim}, which selects from
the surviving entries at every step, retains exactly the set that global
top-$B$ selection over the full context would retain, so irreversible
deletion loses nothing.
For example, both
\begin{equation}
    b_{ij} = c_i-c_j
\end{equation}
and a fixed positional slope
\begin{equation}
    b_{ij}=-\alpha(i-j)
           = -\alpha i + \alpha j
\end{equation}
reduce, after removal of the row constant, to a key-only score.
By contrast, a key-specific decay rate,
\begin{equation}
    b_{ij}
    =
    s_j
    +
    \frac{i-j}{\mathrm{ref}}\log\beta_j,
\end{equation}
cannot remove its $i$ dependence because the coefficient of $i$ itself
depends on $j$.  Its ranking therefore evolves with time, which is useful
for forgetting but conflicts with irreversible eviction: an entry removed
at one step may have become competitive at a later step had it remained
available.  This is the case of our Equation~\ref{eq:retention}, so the
equivalence above does not hold for it.

The policy itself is unchanged.  At every step we still retain the current
top-$B$ of the surviving entries, exactly as in Section~\ref{sec:prelim};
what changes is that this is no longer the same set as top-$B$ over the
full context.  An entry that leaves the retained set is deleted and never
restored, so if the entries kept in its place decay faster and later fall
below it, the cache keeps them and not it.  Because
Equation~\ref{eq:retention} is affine in $i$, the scores of any two
entries cross at most once, so no pair can exchange places twice and an
entry cannot lose to a given rival, regain the lead, and lose again.  This
constrains the individual comparisons but does not prevent an evicted
entry from re-entering the true top-$B$ through crossings with different
rivals, since that boundary is an order statistic over all surviving
entries rather than a single comparison.  Irreversible deletion is
therefore a genuine approximation under Equation~\ref{eq:retention}, and
Section~\ref{sec:method:decay} reports what it costs.

\paragraph{Computational foldability.}
A second requirement is that the bias remain inexpensive to evaluate.
A key-only scalar can be stored once per cached entry.
A key-specific linear decay still has constant fold width:
\begin{equation}
    b_{ij}
    =
    \underbrace{
    \begin{bmatrix}1 & i\end{bmatrix}
    }_{u_i^\top}
    \underbrace{
    \begin{bmatrix}
        s_j-\frac{j}{\mathrm{ref}}\log\beta_j \\
        \frac{1}{\mathrm{ref}}\log\beta_j
    \end{bmatrix}
    }_{v_j}.
\end{equation}
In contrast, cumulative pairwise suppression of the form
$\sum_{j<k<i}S_{kj}$ depends on interactions between a candidate memory
$j$ and the intervening sequence.  In general this object does not admit
a low-width factorization, and therefore requires substantially richer
state or explicit pairwise bias computation.
Our gate sits at the second of these three levels.  Width two is what
keeps the cost of the retention rate confined to bookkeeping: each cached
entry stores $m_j$ and $\log\beta_j$ instead of a single scalar, and a
score at any later position follows from the pair without rescoring the
hidden state, which is what allows one scorer call to serve a batch whose
sequences sit at different absolute positions.

\paragraph{Forced horizons.}
Finally, an age-dependent bias determines whether sufficiently old
memories must eventually disappear regardless of their content.
Suppose the content-dependent score difference is bounded by
$\Delta_{\max}$.  If the relative forgetting penalty between an old and a
new entry grows without bound with their age difference, then beyond some
distance it necessarily dominates $\Delta_{\max}$:
\begin{equation}
    |b_{ij_{\mathrm{old}}}-b_{ij_{\mathrm{new}}}|
    >
    \Delta_{\max}.
\end{equation}
The old entry is then excluded irrespective of its semantic importance,
implicitly imposing a finite horizon.
A per-key gate has no such forced horizon, while a per-key lifetime model
can avoid one only if it is allowed to assign a zero decay rate to entries
that should persist.
Our parameterization is of the second kind: the constraint
$\log\beta_j\le0$ in Equation~\ref{eq:retention} admits zero decay, so no
horizon is imposed by construction and an entry the scorer judges
important can be given an unbounded lifetime.  Our initialization,
however, deliberately starts the lifetime branch away from zero
(Appendix~\ref{app:methoddetails}), and we do not measure whether the
trained rates of the entries that survive longest return to it.

\paragraph{Where this leaves our gate.}
Our gate therefore does not occupy the simplest cell of this family. It is
query-independent in the sense that matters for eviction, namely that the
score never depends on the content of a future query, but it is not
query-independent after canonicalization: the retention rate makes it a
key-specific decay of fold width two. It keeps the two properties that
bear on cost, a compact per-entry state and no forced horizon, and gives
up the third, the frozen ranking that would make online eviction exactly
equivalent to global top-$B_h$ selection. The pilot ablation is consistent
with paying something for that: disabling the retention rate raises the
average from $67.60$ to $69.27$ (Table~\ref{tab:method_components}),
against a two-seed spread of $0.28$ on the same 13-task average. We keep
it for the reason given in Section~\ref{sec:method:decay}, a stable and
explicitly parameterized age scale, and note that the frozen-ranking
variant $b_{ij}=\phi(j)$ is the structurally cleaner choice when
irreversibility is the binding constraint.

\paragraph{Why we relax the gate rather than estimate a gradient through a discrete decision.}
Softening the gate is not the only way to keep the scorer in the computation graph. A per-entry keep-or-drop decision can be sampled and differentiated with a straight-through or Gumbel-softmax estimator, which is the route taken by DMC~\citep{dmc} and DMS~\citep{dms}. We relax the gate instead, for two reasons. First, a straight-through estimator substitutes the gradient of the relaxed variable for the gradient of the decision that was actually executed, so the signal it delivers is biased in a way that does not vanish as training proceeds, whereas the soft gate is differentiated exactly at the value used in the forward pass. Second, the decisions in eviction are coupled: under a cache budget, retaining one entry means dropping another, and independent per-entry Bernoulli variables cannot express this without an added penalty that pushes their expected number towards the budget. Equation~\ref{eq:pinned-gate} enforces the same constraint by construction, since the gates of the evictable set sum to $\beta$ exactly at every step. The relaxation also gives every entry a gradient at every query that can see it, rather than only at the positions where a sampled decision happened to differ.

\paragraph{Why we do not regress an external per-key label.}
A related line supervises the score against a label produced by a separate procedure rather than through the model's own objective. KVzip~\citep{kvzip} scores an entry by the attention it receives while the model is prompted to reconstruct its context, which is query-agnostic but costs a second prefill pass and cannot be evaluated during decoding; Fast KVzip~\citep{fastkvzip} and KVzap~\citep{kvzap} regress that score from hidden states, removing the extra pass. This is cheaper and more stable to train than the objective we use, and we do not claim it performs worse at a matched budget. We do claim it optimizes a different quantity.

\subsection{Recovering evicted information.}
\label{sec:recovering-evicted}

\paragraph{From compression to recovery.}
Summarizing a past that is no longer represented token by token is a
long-standing idea. Compressive memories preserve a coarse record of
activations after they leave the exact attention window
\citep{compressive}; gist tokens bottleneck a prompt into a small number
of learned token states \citep{gist}; and recurrent or test-time-trained
states store history in the activations or parameters of a bounded
memory module \citep{ttt,titans}. These approaches primarily differ in
what is stored and how the state is updated. For KV eviction, however,
there is a separate question: \emph{how should the compressed history
be read relative to the exact entries that remain?}

To make this distinction precise, consider a query $q$ and partition
its original KV cache into retained entries $S$ and evicted entries
$E$. Let
\begin{align}
    Z_A &= \sum_{j\in A}\exp(\ell_j), &
    o_A &= \frac{1}{Z_A}\sum_{j\in A}\exp(\ell_j)v_j,
    \qquad A\in\{S,E\}, \\
    \rho &= \frac{Z_E}{Z_S+Z_E},
    & \ell_j &= qk_j^\top/\sqrt{d}.
\end{align}
The dense-attention output can then be written as
\begin{equation}
    o_{\mathrm{full}}
    = (1-\rho)o_S+\rho o_E,
    \qquad
    o_{\mathrm{evict}}=o_S,
    \label{eq:evicted-decomposition}
\end{equation}
and hence
\begin{equation}
    o_{\mathrm{full}}-o_{\mathrm{evict}}
    = \rho\left(o_E-o_S\right).
\end{equation}
Recovering evicted information therefore requires two distinct
quantities: the \emph{content} or direction $o_E$ contributed by the
evicted entries, and their \emph{mass} $\rho$ relative to the retained
cache. A memory with high nominal capacity can still fail if its
readout does not recover this relative mass.

\paragraph{Output-space residual recovery.}
A common design reads a compressed state independently of attention
and fuses the resulting vector with the exact-attention output:
\begin{equation}
    m_E(q)=\operatorname{Read}(q,M_E),
    \qquad
    o_{\mathrm{outer}}
    =\alpha(q)o_S+\beta(q)m_E(q).
    \label{eq:outer-memory}
\end{equation}
Infini-attention accumulates completed segments in a normalized
associative state and combines its readout with segment-local
softmax attention through a learned scalar gate
\citep{infiniattention}. RAttention maintains exact SWA over recent
tokens and a delayed recurrent linear state over tokens outside the
window, but independently RMS-normalizes and adds the two outputs
\citep{rattention}:
\begin{equation}
    o_t
    =
    \operatorname{RMS}(o_t^{\mathrm{swa}})
    +
    \operatorname{RMS}(o_t^{\mathrm{rla}}).
\end{equation}
Tensor Cache writes each FIFO-evicted KV pair into an outer-product
fast-weight state,
\begin{equation}
    A \leftarrow \lambda A+\eta(k_{\mathrm{old}}\otimes v_{\mathrm{old}}),
    \qquad
    m_t=q_tA,
\end{equation}
and fuses it with local attention using
$o_t=o_t^{\mathrm{local}}+\sigma(g)m_t$
\citep{tensorcache}. The gated-memory variant of Titans and related
TTT memories provide a more expressive state, but use the same general
output-space interface \citep{ttt,titans}.

These constructions can learn useful long-term corrections, especially
when trained end to end. Nevertheless, the normalization required by
Eq.~\ref{eq:evicted-decomposition} is not enforced by their
architecture. Even if $m_E(q)=o_E$ exactly, recovering dense attention
requires $\alpha(q)=1-\rho(q)$ and $\beta(q)=\rho(q)$. A fixed learned
gate, an independently normalized branch, or an additive residual does
not obtain these coefficients from $Z_S$ and $Z_E$; it must instead
learn their calibration indirectly in output space. This distinction
is particularly consequential when retrofitting a frozen backbone,
whose original attention layers already assign a precise probabilistic
meaning to their softmax mass.

\paragraph{Attention-space recovery.}
An alternative is to expose the compressed history as additional KV
entries. Figure~\ref{fig:recovery-architectures} contrasts this
readout with the output-space memories above. Native Hybrid Attention maintains recurrent long-term KV
slots and concatenates them with a local window before a single
softmax \citep{nha}. Key-Value Means merges overflow blocks into a
fixed or growing set of compressed KV rows and likewise attends jointly
to the compressed state and the uncompressed block window, with learned
regional temperatures \citep{kvm}. CentroidKV periodically merges
similar cache entries into degree-weighted centroids and corrects each
centroid logit by the logarithm of its cluster degree
\citep{centroidkv}. Unlike output-space fusion, these methods allow
compressed and exact entries to compete directly for attention mass.

Concretely, let $\mathcal{C}_r\subseteq E$ be a content cluster,
$n_r=|\mathcal{C}_r|$, and $(\bar{k}_r,\bar{v}_r)$ its centroid.
The contribution of $E$ can be approximated by
\begin{align}
    \widehat{Z}_E(q)
    &=
    \sum_{r=1}^{R}
    \exp\left(
        q\bar{k}_r^\top/\sqrt{d}+\log n_r
    \right), \\
    \widehat{N}_E(q)
    &=
    \sum_{r=1}^{R}
    \exp\left(
        q\bar{k}_r^\top/\sqrt{d}+\log n_r
    \right)\bar{v}_r .
\end{align}
Appending these centroids to the retained cache gives
\begin{align}
    o_{\mathrm{joint}}
    &=
    \frac{Z_So_S+\widehat{N}_E}
         {Z_S+\widehat{Z}_E} \\
    &=
    (1-\widehat{\rho})o_S
    +\widehat{\rho}\,\widehat{o}_E,
    \qquad
    \widehat{\rho}
    =
    \frac{\widehat{Z}_E}{Z_S+\widehat{Z}_E},
    \quad
    \widehat{o}_E
    =
    \frac{\widehat{N}_E}{\widehat{Z}_E}.
    \label{eq:joint-recovery}
\end{align}
Thus, the centroid values approximate the missing direction
$o_E$, while their logits and cluster-size corrections estimate the
missing mass $\rho$. Both are reconciled by the same softmax that
weights the retained entries. When a query is unrelated to the
compressed history, the centroids receive negligible mass; when it
matches that history, they take probability mass away from the retained
entries rather than adding an independently scaled residual.

\begin{table}[t]
\centering
\caption{
Representative mechanisms for retaining information that is no longer
stored as individual exact KV entries. ``Joint'' denotes a single
softmax over the exact and compressed supports.
}
\label{tab:evicted-recovery}
\small
\setlength{\tabcolsep}{3.5pt}
\begin{tabularx}{\linewidth}{@{}lXXXc@{}}
\toprule
Method &
Exact support &
Compressed history and write policy &
Readout &
Joint \\
\midrule
Infini-attention
& Current segment
& Associative state over completed segments
& Learned convex gate between two outputs
& No \\

RAttention
& Recent SWA window
& Linear state read only for out-of-window tokens
& Independently RMS-normalized sum
& No \\

Tensor Cache
& Recent SWA window
& Fast-weight state updated on FIFO eviction
& Scalar-gated additive residual
& No \\

NHA
& Recent SWA window
& Fixed recurrent KV slots
& Concatenate slots and local KV
& Yes \\

Key-Value Means
& Recent block window
& Overflow tokens merged into compressed KV rows
& Single softmax with regional temperatures
& Yes \\

CentroidKV
& Sink, recent, and unmerged KV
& Similar KV entries merged into degree-weighted centroids
& Single softmax with $\log n_r$ bias
& Yes \\

\textbf{IndexMem++}
& \textbf{Learned retained set}
& \textbf{Content centroids formed only from learned-evicted entries}
& \textbf{Single softmax with $\log n_r$ bias}
& \textbf{Yes} \\
\bottomrule
\end{tabularx}
\end{table}

IndexMem++ belongs to the attention-space family, but differs in the
source of the exact and compressed sets. Local-memory architectures
partition history by age, while cache-merging methods determine the
compressed set primarily from geometric similarity. We first learn a
query-independent redundancy score that determines which entries must
remain individually addressable, and only then summarize the rejected
set $E$ by content. Selection and recovery are therefore decoupled:
the learned gate decides \emph{what must remain exact}, whereas the
centroids approximate the residual contribution of everything else
under a fixed total support budget $|S|+R$. Section~\ref{sec:method}
shows that enlarging an externally read linear or TTT memory provides
little benefit, while moving a comparable summary into the shared
softmax substantially improves recovery. This indicates that, in this
setting, the limiting factor is not merely memory capacity but the
normalization through which that capacity is accessed.

\section{Training-time kernel for gated attention}
\label{app:gatedattn}

For each layer and KV head, the scorer maps the contextualized hidden state of key $j$ directly to a retention score $s_j=f_\theta(h_j)$.
We suppress the head index for readability. If the retention-rate term in Equation~\ref{eq:retention} is enabled, $s_j$ is replaced by its age-dependent value for query $i$. Let $P_i$ denote the always-attended sinks and causal local window, and let $E_i$ contain the remaining visible keys. The gate added to the attention logits is
\begin{equation}
\bar g_{ij}=
\begin{cases}
0, & j\in P_i,\\[2pt]
\displaystyle s_j-\log\!\sum_{\ell\in E_i}\exp(s_\ell)+\log\beta,
& j\in E_i,
\end{cases}
\qquad
o_i=\sum_j\operatorname{softmax}_j\!\left(
\frac{q_i k_j^\top}{\sqrt d}+\bar g_{ij}
\right)v_j.
\label{eq:training-gated-attn}
\end{equation}
This is the training-time form of Equation~\ref{eq:pinned-gate}, written additively in logit space: $\bar g_{ij}=\log g_{ij}$, so protected entries remain unchanged, while the other entries compete for total gate mass $\beta$. The full causal key axis is used during training so that every score receives a gradient.

\paragraph{Why a standard implementation runs out of memory.}
Although $s_j$ is computed only once per key, the effective gate $\bar g_{ij}$ depends on the query row because each query has a different causal local window $P_i$. A direct implementation therefore broadcasts the per-key scores into a pairwise gate matrix and constructs gated logits of shape $(B,H,S_q,S_k)$. At $S_q=S_k=8192$ with 32 query heads, one bf16 matrix occupies about 4~GiB per layer; retaining these activations for backward across 36 layers requires approximately 144~GiB and causes OOM.

Using a standard attention call with an explicit pairwise gate does not avoid this cost. The always-attended mask prevents the gate from being represented as one static per-key bias, and the resulting SDPA call falls back to the math backend, which materializes the full attention matrix. We therefore implement the operation directly as a fused training kernel.

\paragraph{Fused forward and backward.}
Code~\ref{lst:gated-attn-kernel} shows the core implementation. The MLP computes one $s_j$ per key outside the kernel, and a streaming reduction computes the row-wise gate normalizer. The forward kernel loads these values with each key tile, constructs the sink/local mask arithmetically, and combines the normalized gate with an online attention softmax. The backward kernel recomputes the same tiled logits and accumulates the gradient of each $s_j$ over the queries that attended to it; ordinary autograd then propagates this gradient through the MLP.

\definecolor{kernelcomment}{HTML}{667788}
\definecolor{kernelkeyword}{HTML}{2A63D8}
\definecolor{kernelfunc}{HTML}{3478B8}
\definecolor{kernelbuiltin}{HTML}{168AAD}
\definecolor{kernelconstant}{HTML}{4968A8}
\definecolor{kernelstring}{HTML}{5E76B8}

\lstdefinestyle{gatedattnkernel}{
  language=Python,
  basicstyle=\fontsize{7.4}{8.5}\ttfamily\selectfont,
  keywordstyle=\color{kernelkeyword}\bfseries,
  commentstyle=\color{kernelcomment},
  stringstyle=\color{kernelstring},
  showstringspaces=false,
  columns=fullflexible,
  keepspaces=true,
  breaklines=true,
  frame=tb,
  rulecolor=\color{kernelcomment},
  framerule=0.5pt,
  framesep=4pt,
  xleftmargin=2pt,
  xrightmargin=2pt,
  aboveskip=4pt,
  belowskip=3pt,
  captionpos=b,
}

\begin{lstlisting}[
  style=gatedattnkernel,
  caption={Core training-time gated-attention kernel.},
  label={lst:gated-attn-kernel}
]
g = MLP(h)                         # [batch, H_kv, length]
lse = streaming_logsumexp(g, evictable, beta)

# One Triton program owns a query tile and streams key tiles.
def forward(Q, K, V, g, lse):
    row_max = -inf
    row_sum = 0
    out_acc = 0

    for key_tile in key_tiles:
        q = load(Q, query_tile)
        k = load(K, key_tile)
        v = load(V, key_tile)
        g_key = load(g, key_tile)

        logits = q @ k.T / sqrt(d)
        protected = is_sink(key_tile) | is_local(query_tile, key_tile)
        gate = where(protected, 0, g_key - lse[:, None])
        logits = where(causal(query_tile, key_tile), logits + gate, -inf)

        # Online softmax; no pairwise gate or attention matrix is stored.
        new_max = max(row_max, rowmax(logits))
        rescale = exp(row_max - new_max)
        p = exp(logits - new_max[:, None])
        out_acc = out_acc * rescale[:, None] + p @ v
        row_sum = row_sum * rescale + rowsum(p)
        row_max = new_max

    out = out_acc / row_sum[:, None]
    row_lse = row_max + log(row_sum) # saved for backward
    return out, row_lse


def backward(Q, K, V, g, lse, out, row_lse, d_out):
    delta = rowsum(out * d_out)
    dQ, dK, dV, dg, d_lse = zeros()

    for key_tile in key_tiles:
        # Recompute the same gated logits instead of saving O(L^2) activations.
        logits, protected = recompute_logits(Q, K, g, lse, key_tile)
        p = exp(logits - row_lse[:, None])

        atomic_add(dV[key_tile], p.T @ d_out)
        dP = d_out @ V[key_tile].T
        dZ = p * (dP - delta[:, None])

        dQ += dZ @ K[key_tile] / sqrt(d)
        atomic_add(dK[key_tile], dZ.T @ Q / sqrt(d))
        d_gate = where(protected, 0, dZ)
        atomic_add(dg[key_tile], rowsum_over_queries(d_gate))
        d_lse += -rowsum_over_keys(d_gate)

    return dQ, dK, dV, dg, d_lse
\end{lstlisting}

Neither pass materializes the pairwise gate, logits, or attention probabilities. The saved activation memory is therefore $O(S_q+S_k)$ rather than $O(S_qS_k)$, although the arithmetic remains quadratic because every visible key receives an independent training signal. Softmax and gradient accumulation use fp32 for numerical stability.

A causal row can initially contain only always-attended entries, leaving $E_i$ empty and its log-sum-exp undefined. Such rows use an inert gate, which preserves dense attention and prevents NaNs without suppressing any meaningful retention signal.

\section{From the soft gate to hard eviction}
\label{app:hardevict}

At inference, the learned gate is used only to score cache entries, while attention is computed normally over the retained entries. Our implementation uses sparse attention to simulate eviction during prefill, then converts the final prefill state into a physically compressed cache for streaming decode.

\paragraph{Sparse prefill.} During prefill, each query attends only to its sinks, local window, and highest-scoring visible entries. We evaluate the retention score once for the whole prefill and hold that score vector fixed while the mask is built. Against a fixed score vector the candidate pool of successive queries only grows, so the number of entries outranking a given key is non-decreasing and each key is selected by one contiguous interval of query rows. Each key's support is then summarized by the last position at which it survives, and the mask is expressed as a FlexAttention block mask, avoiding a per-query top-$B_h$ support tensor and indirect KV gathers. This reproduces per-row top-$B_h$ selection over that score vector exactly; with the per-key retention rate enabled, holding the vector fixed is an additional prefill approximation, since a query midway through the prefill would otherwise rank entries by their scores at its own position. The full KV cache is retained until prefill finishes, so this stage reduces attention computation but not cache storage.

\paragraph{Streaming eviction.} After prefill, we copy the final row's retained entries into the compressed cache and release the dense cache. During decoding, each new token first enters the local window; the entry leaving the window then competes with the weakest entry in the retained pool, so at most one cache slot is replaced per head and step and the cache size remains fixed. This update is exactly equivalent to global top-$B_h$ selection when the ordering is fixed; with learned per-key decay, it is an irreversible online approximation because deleted entries cannot be reconsidered after their scores cross.

\paragraph{Ragged paged cache.} The head allocator assigns a different budget $B_h$ to each KV head, so padding all heads to the largest budget would waste much of the saved memory. We instead store each $(\text{layer},\text{sequence},\text{KV head})$ row in its own pages. Decode then performs dense paged GQA directly over the retained entries, without a mask or gather, and both cache storage and historical KV reads scale with $\sum_h B_h$ rather than the original context length, as reflected in Figure~\ref{fig:exp-decode-throughput}.

\section{Efficiency}
\label{app:efficiency}

\paragraph{Measurement setup.} Figure~\ref{fig:exp-decode-throughput}, summarized in Section~\ref{sec:exp:efficiency}, reports aggregate batch decode throughput (token/s) on one NVIDIA H20 GPU for Qwen3-4B and Qwen3-8B. Input lengths are 8K, 16K, 32K, 64K, 128K, and 256K (where 1K denotes 1,024 tokens), with batch sizes $B\in\{1,4,16,32\}$, giving 48 configurations. Every configuration decodes 256 steps in BF16, and prefill, initial compression, graph capture, and one-time setup are excluded from the timing. IndexMem++ decodes through a captured CUDA Graph on deterministic pseudo-random token inputs and we report the fastest of three repetitions; its average per-KV-head budget is 2,048 entries, inclusive of the 64 CMP slots, the four sink tokens, and the 128-token local window, under the mass-based head allocation of Table~\ref{tab:exp:inference-config}. The Full-KV* reference decodes eagerly with FlashAttention-2 on GovReport inputs, aggregated over 15 measurements. The star marks this difference in decode implementation: part of any gap between the two curves reflects graph replay rather than the retained-cache budget, so we read the figure for how each curve scales with input length and treat cross-curve ratios as indicative only.

\paragraph{Context-length scaling.} A fixed retained KV budget largely decouples decode throughput from the original input length. From 8K to 256K, IndexMem++ throughput at $B=1$ changes from 53.95 to 52.18 token/s for Qwen3-4B and from 48.38 to 46.51 token/s for Qwen3-8B, while the Full-KV* reference measurements fall from 36.85 to 8.42 and from 33.88 to 8.21, respectively. At $B=4$, IndexMem++ changes from 173.69 to 174.69 and from 166.96 to 155.95 token/s for the two models. The $B=16$ and $B=32$ curves decline more, but IndexMem++ remains substantially less sensitive to the original context length than the Full-KV* reference curves. This difference is consistent with the execution model in Appendix~\ref{app:hardevict}: continuous eviction bounds the historical KV read by IndexMem++ to the retained cache budget, whereas Full-KV reads a historical cache that grows with sequence length.

\paragraph{Batch-size scaling.} Within the IndexMem++ measurements, throughput increases substantially with batch size while each sequence retains the same bounded historical-cache budget: at 16K it rises from 53.13 to 685.72 token/s for Qwen3-4B and from 47.19 to 630.07 token/s for Qwen3-8B as $B$ goes from 1 to 32. The Full-KV* reference is comparatively flat in batch size at this length, since each added sequence brings its own growing history to read. The two effects compound with the implementation difference noted above, so we do not attribute a specific ratio to the budget alone. In our implementation, scoring and cache maintenance are also batched across sequences and layers.

\paragraph{Benchmark coverage and cache footprint.} Figure~\ref{fig:exp-decode-throughput} reports IndexMem++ decode throughput for all 48 model--length--batch configurations. The Full-KV* reference contains 32 measured configurations and 16 OOM cases. For both models, Full-KV* fails from 32K onward at $B=32$, from 64K onward at $B=16$, and at 256K for $B=4$, while $B=1$ succeeds through 256K. The ragged paged cache in Appendix~\ref{app:hardevict} allocates physical KV storage according to fixed per-head retained budgets, so retained-cache storage scales with $\sum_h B_h$ per layer and sequence rather than with the original context length. Inputs of 64K--256K exceed the 40,960-position configuration used for quality evaluation and serve as throughput and memory stress tests, not long-context accuracy results.

\section{Details deferred from the method}
\label{app:methoddetails}


\subsection{Both halves of the always-attended construction are required.}
Normalizing all gates without exempting any entry is ineffective because the row-wise normalizer adds the same constant to every attention logit and therefore cancels inside the softmax. Exempting sinks and the local window alone is also insufficient: the remaining gates can all approach $1$ and exactly recover dense attention. Equation~\ref{eq:pinned-gate} combines the two requirements. Protected entries remain fixed at one, while the evictable entries must share a total gate mass $\beta$; they therefore cannot all equal one and must compete with one another. This removes the dense no-op solution and turns $\beta$ into an exact training-time budget rather than a target imposed through a penalty.

\paragraph{What the scorer learns at the constant point.}
The construction does not forbid a constant score, and it does not need to. Substituting $s_j\equiv c$ into Equation~\ref{eq:pinned-gate} gives $g_{ij}=\beta/|E_i|$ regardless of $c$: a single policy that attenuates every evictable entry equally and is dense only if $\beta=|E_i|$. Because that policy is not dense, the loss gradient there is nonzero. Writing $r_{ij}=\langle\partial\mathcal L/\partial o_i,\,v_j\rangle$ for the sensitivity of the loss to entry $j$ through the output of query $i$, and $\bar r_i=\sum_k a_{ik}r_{ik}$ for its attention-weighted mean, the gradient with respect to a score is
\begin{equation}
\frac{\partial\mathcal L}{\partial s_m}
=\sum_{i:\,m\in E_i}\Big(D_{im}-p^{(s)}_{im}\!\!\sum_{j\in E_i}\!D_{ij}\Big),
\qquad D_{ij}=a_{ij}\big(r_{ij}-\bar r_i\big),
\qquad p^{(s)}_{im}=\frac{\exp(s_m)}{\sum_{\ell\in E_i}\exp(s_\ell)} .
\label{eq:gate-gradient}
\end{equation}
Each score thus receives the advantage of its own entry over the current evictable average: an entry contributing more than average gains gate mass, and it gains it at the expense of the others. Competition for the limited mass, rather than the unavailability of a constant score, is what drives the scorer to rank entries.

Two further consequences are worth stating. First, the separation from dense attention is graded rather than binary: the constant solution attenuates evictable entries by $\beta/|E_i|$ relative to the protected ones, so the pressure away from the no-op grows with $\log(|E_i|/\beta)$ and disappears as $\beta\to|E_i|$. The requirement is therefore $\beta\ll|E_i|$ rather than merely $\beta<|E_i|$; in the pilot setting $\beta/|E_i|\approx 256/8064$, roughly $3\%$. This is also why $\beta$ is held fixed rather than scaled with the sequence length, since a $\beta$ growing with $|E_i|$ would weaken the mechanism at exactly the lengths that need it. Second, summing Equation~\ref{eq:gate-gradient} over a uniform shift of all scores within a head gives exactly zero, because $\sum_{m\in E_i}(\delta_{jm}-p^{(s)}_{im})=0$. The absolute score level is therefore unidentifiable, which is the same degeneracy that prevents raw scores from being compared across heads and motivates the retained-mass allocator.

\subsection{Loss design.}
Forward $\mathrm{KL}(p^{\mathrm{dense}}\|p^{\mathrm{gated}})$ is mass-covering, or mean-seeking: it encourages the gated model to cover the full teacher distribution, including its lower-probability alternatives. Reverse $\mathrm{KL}(p^{\mathrm{gated}}\|p^{\mathrm{dense}})$ is mode-seeking: it concentrates the student on the teacher's highest-probability modes and penalizes probability assigned outside them.

\paragraph{The two directions agree to first order.} Writing $q=p^{\mathrm{gated}}$ and $p=p^{\mathrm{dense}}$ for one position and differentiating with respect to the gated logits $z$,
\begin{equation}
\frac{\partial\,\mathrm{KL}(p\|q)}{\partial z_k}=q_k-p_k,
\qquad
\frac{\partial\,\mathrm{KL}(q\|p)}{\partial z_k}=q_k\big(r_k-\bar r\big),
\qquad
r_k=\log\frac{q_k}{p_k},
\quad
\bar r=\sum_j q_jr_j .
\label{eq:kl-directions}
\end{equation}
The forward direction reproduces the first term of Equation~\ref{eq:ce-decomposition} exactly. The reverse direction is not a reweighting of it, since components with $0<r_k<\bar r$ receive a gradient of the opposite sign, but the two coincide near the matching point: substituting $q=p+\epsilon$ gives $r_k=\epsilon_k/p_k+O(\|\epsilon\|^2)$ and $\bar r=O(\|\epsilon\|^2)$, because $\sum_k\epsilon_k=0$, so $q_k(r_k-\bar r)=\epsilon_k+O(\|\epsilon\|^2)$. Both directions therefore share the same leading-order gradient and differ only in higher-order terms, which is where the mass-covering and mode-seeking behaviors originate. Neither direction contains the frozen-backbone term of Equation~\ref{eq:ce-decomposition}, so the choice between them is not a question of which error is isolated and we make it empirically.

We also add cross-entropy with weight $0.1$, chosen to place its contribution on approximately the same numerical scale as the KL term; this changes the average from $67.60$ to $67.33$. Adding this term reintroduces the frozen-backbone error $p^{\mathrm{dense}}_t-e_{y_t}$ into the gradient, and the result does not improve, so the final recipe uses no cross-entropy term and the language-modeling target provides no additional benefit in this setting.

We try LongCE because it emphasizes tokens whose prediction benefits from long context, which should focus training on positions most affected by KV eviction. However, LongCE only reweights the cross-entropy gradient: it scales the error caused by eviction and the irreducible error of the frozen dense model together, rather than removing the latter as distillation does. It therefore improves CE only from $51.50$ to $53.01$, remaining far below reverse KL at $67.60$ (Table~\ref{tab:method_components}).

In future work we plan to do on-policy KL distillation, where training prefixes are generated by the current hard-eviction policy rather than by dense or teacher-forced trajectories. This would expose the gate to the state distribution and accumulated errors it encounters at inference, thereby reducing the train--inference mismatch. It is also attractive for reinforcement-learning pipelines, whose rollouts maintain many long-lived KV caches: applying eviction during policy generation bounds the rollout cache by the retained budget and can reduce the memory and bandwidth cost of sampling, although dense teacher or reference scoring may still require a separate full-cache pass.

\subsection{Heads budget design.}
Adding a constant $c$ to all retention scores of one head shifts its normalizer by the same amount, leaving Equation~\ref{eq:pinned-gate} and the within-head top-$k$ ordering unchanged. The absolute score level is therefore not identifiable across heads, so pooling raw scores across heads cannot determine their relative budgets. Retained attention mass is invariant to this shift and provides a comparable measure of demand. For the final prefill query, we order each head's evictable entries by its retention score and define $\rho_h(k)$ as the attention mass retained by the first $k$ entries together with the protected sinks and window; query heads sharing one KV head are averaged. We choose the smallest $k_h$ for which $\rho_h(k_h)\geq\rho^\star$ and adjust the shared target $\rho^\star$ until the total budget is met. Any integer remainder is assigned by marginal mass: an extra slot goes to the head whose next entry adds the most attention mass, or is removed from the head whose last entry contributes the least. This is a discrete water-filling allocation that preserves the layer budget exactly. Under a uniform split, the retained-mass range across heads is $0.268$ on average and reaches $0.606$ in layer 4, leaving substantial room for reallocation.

Both controls preserve exactly the same total number of KV entries. The permuted control shuffles the online allocator's budget vector across heads, preserving its ragged shape while breaking the correspondence between each head and its measured demand. Its score falls to $57.14$, below the uniform allocation at $58.86$, showing that unequal budgets alone do not explain the improvement. The offline control averages per-head allocations over a calibration set and uses the resulting fixed layer--head table for every input. It reaches $62.33$, recovering part of the gain without online measurement and showing that some head specialization is stable across inputs. Its remaining gap to the context-dependent allocator at $67.60$ shows that the appropriate allocation also depends on the current context (Table~\ref{tab:method_components}).

Code~\ref{lst:head-budget-mass} summarizes the online allocation.

\begin{lstlisting}[
  style=gatedattnkernel,
  caption={Core retained-mass head-budget allocation.},
  label={lst:head-budget-mass},
  emph={mass_curve,allocate_by_mass,head_budgets},
  emphstyle=\color{kernelfunc}\bfseries,
  emph={[2]softmax,argsort,cumsum,gather,mean,argmin,argmax},
  emphstyle={[2]\color{kernelbuiltin}},
  emph={[3]True,False,None,inf},
  emphstyle={[3]\color{kernelconstant}\bfseries}
]
# Scores rank entries within a head; attention mass allocates across heads.
def mass_curve(Q, K, score, ref_row, sinks, window):
    logits = Q[:, ref_row] @ K.T / sqrt(d)
    prob = softmax(mask_future(logits, ref_row), dim=-1)
    prob = mean_over_gqa_group(prob)       # [H_kv, length]

    pinned = is_sink(sinks) | is_local(ref_row, window)
    pool = is_causal(ref_row) & ~pinned
    pool_size = sum(pool)
    order = argsort(where(pool, score, -inf), descending=True, stable=True)

    pinned_mass = rowsum(where(pinned, prob, 0))
    ranked_mass = gather(prob, order[:, :pool_size])
    return pinned_mass[:, None] + cumsum(ranked_mass, dim=-1)


def allocate_by_mass(curve, total_slots):
    H_kv, pool_size = curve.shape
    # Fewest slots with retained mass at least rho.
    def counts(rho):
        reached = curve >= rho
        return where(any(reached, dim=-1), first_true(reached) + 1, pool_size)

    # Find a common retained-mass target.
    lo, hi = 0.0, 1.0
    for _ in range(60):
        rho = (lo + hi) / 2
        if sum(counts(rho)) > total_slots: hi = rho
        else: lo = rho
    k = counts(lo)

    # Integer correction keeps the layer budget exact.
    while sum(k) < total_slots: gain = mass_of_next_entry(curve, k), k[argmax(gain)] += 1
    while sum(k) > total_slots: loss = mass_of_last_entry(curve, k), k[argmin(loss)] -= 1
    return k


def head_budgets(Q, K, score, topk, sinks, window):
    curve = mass_curve(Q, K, score, last_prefill_row, sinks, window)
    H_kv = curve.shape[0]
    evictable = topk - sinks - window
    k = allocate_by_mass(curve, H_kv * evictable)
    return k + sinks + window
\end{lstlisting}

\subsection{The three history scorers, and why each is a single-variable test.}
All three variants retain the token-local MLP and add one history branch before its output nonlinearity. Writing $x_j=\operatorname{Norm}(h_j)$ and $a_j$ for the history feature, the shared form is $z_j=\operatorname{GELU}(\operatorname{Norm}(W_{\mathrm{in}}x_j+W_a\operatorname{Norm}(a_j)))$, followed by the same per-head score projection as the reference scorer. We initialize $W_a$ to zero, so every variant begins exactly from the token-local MLP and differs only in the construction of $a_j$. Figure~\ref{fig:gate-architectures} shows the five scorers side by side, including the linear readout and the token-local MLP used as references.

\paragraph{Causal convolution.} We project the normalized hidden states to $u_j=W_cx_j$ and apply a width-eight depthwise causal convolution over $u_{j-8},\ldots,u_{j-1}$. The current token is excluded because its information is already carried by the MLP branch. Each channel has its own signed temporal filter, allowing the scorer to compare a token with its recent neighborhood while requiring only the last eight projected states during decoding.

\paragraph{Gated recurrent state.} We maintain a per-channel state $S_j=g_j\odot S_{j-1}+(1-g_j)\odot u_j$, where $u_j=W_ux_j$ and $g_j=\sigma(W_gx_j)$. The score of token $j$ reads $S_{j-1}$ rather than $S_j$, so this branch contains only information from preceding tokens. The input-dependent gate lets different channels retain or replace information over different time scales. During training the recurrence is evaluated in fp32 with an associative scan, while decoding carries one fixed-size state vector per layer.

\paragraph{Prefix attention.} We give the scorer a separate single-head causal attention with $q_j=W_qx_j$, $k_i=W_kx_i$, $v_i=W_vx_i$, and $a_j=\operatorname{softmax}(q_jK_{\leq j}^{\top}/\sqrt D)V_{\leq j}$. This is still query-independent for eviction: $q_j$ is formed from the hidden state of key token $j$, rather than from a later model query, so the score is fixed when the key is created. It provides access to the entire prefix, but incremental scoring requires retaining the scorer's own K/V cache and attending over a history that grows with the sequence.

The convolution, recurrent, and prefix-attention scorers use 16.8M, 24.1M, and 19.5M parameters, respectively, compared with 7.5M for the token-local MLP, yet change the RULER average by only $-0.78$, $+1.16$, and $-0.97$ points (Table~\ref{tab:method_components}). A Fast-KVzip-style adaptive-reference head is likewise within $0.35$ points. We therefore conclude that sufficient scorer capacity is necessary at this model scale, but explicitly re-reading the history provides little benefit because $h_j$ is already contextualized.

\subsection{Retention-rate}
Equation~\ref{eq:retention} uses a fixed age scale $\tau$ rather than the current sequence length. Without normalization, $\partial s_j(i)/\partial\log\beta_j=i-j$ can be $10^4$ of magnitude larger than $\partial s_j(i)/\partial m_j=1$ at long context lengths, making the lifetime head effectively train at a much larger learning rate than the magnitude head. Dividing by a fixed $\tau$ keeps the two derivatives comparable and gives $\log\beta_j$ a consistent interpretation as the score decrease over $\tau$ tokens. Using the live sequence length instead would continually rescale the age contribution of every cached key as the context grows, introducing changes in which entries are retained that are caused by the changing denominator rather than by elapsed age; in a positional-score diagnostic, the fixed scale produced no re-entries over 1,500 steps, whereas live-length normalization produced 27.

\paragraph{Retention-rate initialization.} We initialize $\log\beta_j$ so that its change over one reference interval is comparable to the standard deviation of the magnitude score. This differs from TrimKV, where decay starts near zero and is driven by a separate retention hinge. Our fixed-mass normalizer enforces a budget but does not itself encourage shorter lifetimes, so a near-zero initialization could leave the lifetime branch inactive and reduce the model to the content-only scorer. The retention rate does not provide a consistent accuracy gain: its effect is approximately neutral under a uniform budget and reverses under the head-budget allocator. We therefore retain it for a stable, explicitly parameterized age scale rather than as an accuracy-improving component.

\section{Pilot ablation protocol}
\label{app:ablation}
\paragraph{Protocol.}
Every arm in Tables~\ref{tab:method_ladder}, \ref{tab:method_joint} uses Qwen3-0.6B and RULER at an 8K context length. Eviction evaluation retains $25\%$ of the context. Unless explicitly stated otherwise, evaluation uses seed 42. Each gate is trained for 100 steps at 8K with a gate-mass budget of 256, peak learning rate $10^{-3}$, and data seed 1000.

\paragraph{Two tasks carry no signal at this model size.}
At this scale, \texttt{cwe} remains between 0.00 and 0.65 for every arm, indicating that Qwen3-0.6B cannot solve the task under this protocol. \texttt{multikey3} ranges from 0.00 to 47.83 across arms, but two seeds of the same reference configuration score 26.09 and 25.40; its apparent between-arm variation is therefore not cleanly separable from sampling variation. Both tasks remain in Avg. because they belong to the standard 13-task RULER suite, but no claim in Section~\ref{sec:method} relies on either individual column.

\section{IndexMem++ implementation settings}
\label{app:exp:protocol}

This section gives the complete training recipe and the inference-time configuration promised by Section~\ref{sec:exp:setup}.

\paragraph{Training data.}
We train on LongMino, using its 256K-filtered split in pre-tokenized form and drawing from the \texttt{2e15}, \texttt{2e16}, \texttt{synth\_cwe}, and \texttt{synth\_rex} subsets. The DMA baseline uses the same mixture. Its value-derived, query-independent score is trained with the LM loss as an in-softmax gate and used for hard eviction. DMA is trained for 300 optimizer steps at 8K only, with global batch 8 and the backbone frozen. Documents are drawn at random with a shuffle buffer of 64 and a data seed of 1000.

\paragraph{Length curriculum and optimization.}
The gate is trained under a length curriculum of 300 steps at 8K, 300 steps at 16K, with global batch of 8. We use AdamW with a peak learning rate of $10^{-3}$ decaying to $5\times10^{-6}$, warming up over the first $10\%$ of steps and holding the peak for the following $60\%$. Only the gate parameters are placed in the optimizer; the backbone, the embeddings, and the LM head remain frozen throughout, and Section~\ref{sec:method} reports what unfreezing them costs.

\paragraph{Gate configuration.}
The scorer is a two-layer MLP applied to the hidden state $h_j$ with a middle width of 256, emitting one score per KV head. The per-key retention rate of Equation~\ref{eq:retention} is enabled. Four attention sinks and a 128-token causal window are always attended; these entries are exempt from eviction and from the gate normalizer. The gate mass in Equation~\ref{eq:pinned-gate} is $\beta=256$.

\paragraph{Inference-time components.}
Two components are applied only at inference and introduce no trained parameters. The head-budget allocator redistributes a fixed layer-wide support budget across KV heads by retained attention mass. Let $B_h$ denote the total support assigned to KV head $h$. Sink entries, local-window entries, and the centroid slots are funded within $B_h$, rather than added on top. Under mass allocation, the layer-wide budget satisfies $\sum_h B_h = H_{\mathrm{KV}}K$, where $K$ is the average per-KV-head support used for the reported compression setting.

The allocator is subject to a per-head floor, which we set to $K/4$ rather than to an absolute number of entries. A floor fixed in absolute terms would interact with the budget: since the layer total is exactly $H_{\mathrm{KV}}K$ and every head must reach the floor, a floor equal to $K$ would force every head to exactly $K$ and reduce the allocator to a uniform split, and a floor close to $K$ would leave it little room to reallocate. Tying the floor to $K$ keeps the same fraction of the budget available for redistribution at every compression setting, so the allocator behaves consistently across the budgets reported in this paper. At the RULER setting $K=2048$ this gives a floor of 512 entries.

\paragraph{Per-benchmark inference configuration.}
Table~\ref{tab:exp:inference-config} lists the inference-time configuration of IndexMem++ on each benchmark. The trained gate and the always-attended regions are the same everywhere, and the budget is counted the same way. RULER and LongBench differ only in the budget $K$ implied by the compression ratio. On Math500 and AIME25 the prompt is short and the cache is filled almost entirely during generation, so the mass allocator, which reads the attention at the end of prefill, has almost nothing to work from. There IndexMem++ uses a static head table fitted offline for each budget, with the same $K/4$ floor, in place of the per-document allocator, and no centroid slots. Each table is fitted once by running the per-document allocator on 16 LongMino documents of 8K tokens and averaging the resulting per-head budgets.

\begin{table}[!htbp]
\centering
\caption{\textbf{IndexMem++ inference configuration per benchmark.} One trained gate is used throughout; the budget $K$ is the average per-KV-head total support and is inclusive of every protected entry and centroid slot listed here.}
\label{tab:exp:inference-config}
\begingroup\fontsize{8.7}{10.3}\selectfont
\setlength{\tabcolsep}{4pt}\renewcommand{\arraystretch}{1.18}
\begin{tabular}{@{}lccccc@{}}
\toprule
Benchmark & Budget $K$ & Allocator & Floor & Sink / local & CMP \\
\midrule
RULER & $\lfloor(1-c)N\rfloor$ & mass & $K/4$ & 4 / 128 & 64 \\
LongBench & $\lfloor(1-c)N\rfloor$ & mass & $K/4$ & 4 / 128 & 64 \\
Math500 & 512, 1{,}024 & static & $K/4$ & 4 / 128 & 0 \\
AIME25 & 1{,}024, 2{,}048 & static & $K/4$ & 4 / 128 & 0 \\
\bottomrule
\end{tabular}
\endgroup
\end{table}

\section{Additional Long-Context Results}
\label{app:exp:longcontext}

\subsection{Complete RULER results}
\label{app:exp:ruler}
The tables below report Full-KV and the displayed methods at compression ratios 50\% and 75\%, across all 13 RULER tasks. The average includes all 13 tasks.
\begin{table}[!htbp]
\centering
\caption{\textbf{Complete RULER scores: Qwen3-4B, 8K.} Compression ratios 50\% and 75\%.}
\label{tab:exp:app-ruler-4b-8192}
\begingroup\fontsize{7.6}{9.4}\selectfont
\setlength{\tabcolsep}{1.2pt}\renewcommand{\arraystretch}{1.18}
\setlength{\ExpMethodWidth}{0.159\linewidth}\setlength{\ExpNumWidth}{\dimexpr(\linewidth-\ExpMethodWidth-28\tabcolsep)/14\relax}\begin{tabular}{@{}p{\ExpMethodWidth}*{14}{>{\raggedleft\arraybackslash}p{\ExpNumWidth}}@{}}
\toprule
Method & S1 & S2 & S3 & MK1 & MK2 & MK3 & MV & MQ & VT & CWE & FWE & QA1 & QA2 & Avg. \\
\midrule
\rowcolor{ExpReference}
Full-KV & 100.00 & 100.00 & 100.00 & 99.80 & 99.80 & 99.80 & 99.85 & 99.70 & 99.96 & 84.56 & 86.27 & 75.20 & 56.40 & 92.41 \\
\midrule
\multicolumn{15}{l}{\textbf{Compression ratio: 50\%}} \\
SnapKV & 85.20 & 89.00 & 5.60 & 43.60 & 31.60 & 28.80 & 43.00 & 41.05 & 78.12 & 77.10 & \textbf{93.27} & 45.80 & 43.00 & 54.24 \\
KeyDiff & \textbf{100.00} & \textbf{100.00} & 98.80 & 91.80 & 43.00 & 11.00 & 98.85 & 95.80 & \textbf{99.96} & \textbf{80.98} & 82.80 & 34.00 & 21.40 & 73.72 \\
DMA & 99.80 & 84.80 & 29.80 & 80.60 & 61.60 & 0.60 & 81.85 & 88.60 & 96.20 & 56.44 & 85.80 & 70.80 & 49.40 & 68.18 \\
TrimKV & \textbf{100.00} & 99.60 & 99.80 & \textbf{99.80} & 99.00 & 42.40 & 99.65 & \textbf{99.95} & \textbf{99.96} & 80.88 & 77.73 & 67.20 & 57.00 & 86.38 \\
\rowcolor{ExpOurs}
\textbf{IndexMem++} & \textbf{100.00} & \textbf{100.00} & \textbf{100.00} & 99.60 & \textbf{99.60} & \textbf{99.40} & \textbf{99.80} & 99.90 & \textbf{99.96} & 75.20 & 86.27 & \textbf{75.40} & \textbf{57.60} & \textbf{91.75} \\
\midrule
\multicolumn{15}{l}{\textbf{Compression ratio: 75\%}} \\
SnapKV & 70.20 & 22.20 & 2.40 & 15.40 & 8.00 & 3.40 & 14.05 & 13.85 & 38.36 & 62.18 & \textbf{89.00} & 29.60 & 32.40 & 30.85 \\
KeyDiff & \textbf{100.00} & 88.60 & 92.00 & 74.60 & 4.60 & 0.00 & 89.90 & 89.50 & \textbf{100.00} & 69.14 & 74.07 & 15.00 & 12.60 & 62.31 \\
DMA & 99.00 & 24.20 & 2.40 & 27.40 & 5.40 & 0.00 & 14.30 & 20.70 & 79.60 & 48.16 & 71.47 & 47.80 & 32.00 & 36.34 \\
TrimKV & \textbf{100.00} & 98.80 & \textbf{92.60} & 95.00 & 48.40 & 10.80 & 91.55 & 95.60 & 99.96 & \textbf{77.00} & 71.60 & 49.60 & 49.60 & 75.42 \\
\rowcolor{ExpOurs}
\textbf{IndexMem++} & \textbf{100.00} & \textbf{99.20} & 91.20 & \textbf{97.80} & \textbf{94.80} & \textbf{60.40} & \textbf{98.40} & \textbf{99.20} & \textbf{100.00} & 44.28 & 84.07 & \textbf{68.00} & \textbf{53.20} & \textbf{83.89} \\
\bottomrule\end{tabular}
\endgroup
\par\vspace{3pt}\begin{minipage}{\linewidth}\fontsize{8}{9.3}\selectfont Task abbreviations follow Table~\ref{tab:exp-ruler-tasks}.\end{minipage}
\end{table}

\begin{table}[!htbp]
\centering
\caption{\textbf{Complete RULER scores: Qwen3-4B, 16K.} Compression ratios 50\% and 75\%.}
\label{tab:exp:app-ruler-4b-16384}
\begingroup\fontsize{7.6}{9.4}\selectfont
\setlength{\tabcolsep}{1.2pt}\renewcommand{\arraystretch}{1.18}
\setlength{\ExpMethodWidth}{0.159\linewidth}\setlength{\ExpNumWidth}{\dimexpr(\linewidth-\ExpMethodWidth-28\tabcolsep)/14\relax}\begin{tabular}{@{}p{\ExpMethodWidth}*{14}{>{\raggedleft\arraybackslash}p{\ExpNumWidth}}@{}}
\toprule
Method & S1 & S2 & S3 & MK1 & MK2 & MK3 & MV & MQ & VT & CWE & FWE & QA1 & QA2 & Avg. \\
\midrule
\rowcolor{ExpReference}
Full-KV & 100.00 & 100.00 & 100.00 & 99.40 & 99.00 & 98.80 & 99.95 & 99.95 & 99.80 & 69.82 & 97.00 & 72.80 & 54.80 & 91.64 \\
\midrule
\multicolumn{15}{l}{\textbf{Compression ratio: 50\%}} \\
SnapKV & 92.40 & 93.80 & 13.40 & 69.20 & 31.40 & 21.40 & 60.10 & 74.35 & 94.36 & 66.28 & 98.27 & 43.00 & 43.00 & 61.61 \\
KeyDiff & \textbf{100.00} & 99.80 & 99.40 & 75.00 & 27.80 & 3.80 & 93.60 & 95.80 & \textbf{99.88} & \textbf{70.16} & 90.73 & 29.00 & 25.00 & 70.00 \\
DMA & \textbf{100.00} & 83.40 & 32.60 & 77.20 & 51.60 & 0.20 & 79.35 & 83.75 & 92.56 & 33.70 & 94.07 & 65.60 & 47.00 & 64.69 \\
TrimKV & \textbf{100.00} & \textbf{100.00} & \textbf{100.00} & 98.60 & 97.40 & 40.40 & 99.15 & \textbf{99.90} & 99.80 & 63.98 & \textbf{99.13} & 71.60 & 53.20 & 86.40 \\
\rowcolor{ExpOurs}
\textbf{IndexMem++} & \textbf{100.00} & \textbf{100.00} & \textbf{100.00} & \textbf{99.00} & \textbf{98.40} & \textbf{85.20} & \textbf{99.85} & \textbf{99.90} & 99.68 & 52.98 & 97.47 & \textbf{72.60} & \textbf{54.20} & \textbf{89.18} \\
\midrule
\multicolumn{15}{l}{\textbf{Compression ratio: 75\%}} \\
SnapKV & 72.60 & 39.00 & 4.40 & 19.60 & 5.60 & 5.20 & 14.60 & 20.45 & 73.40 & 51.92 & 97.00 & 30.80 & 30.80 & 35.80 \\
KeyDiff & \textbf{100.00} & 87.00 & 92.60 & 65.20 & 2.20 & 0.00 & 91.30 & 89.10 & 99.48 & \textbf{65.74} & 83.87 & 14.00 & 15.40 & 61.99 \\
DMA & 98.60 & 20.40 & 2.40 & 17.60 & 2.20 & 0.00 & 14.00 & 15.30 & 76.96 & 30.26 & 77.20 & 45.00 & 33.00 & 33.30 \\
TrimKV & \textbf{100.00} & 97.40 & 92.80 & 88.40 & \textbf{42.40} & 13.80 & 93.60 & 95.55 & 98.60 & 60.34 & 96.67 & 65.20 & 47.00 & 76.29 \\
\rowcolor{ExpOurs}
\textbf{IndexMem++} & \textbf{100.00} & \textbf{100.00} & \textbf{98.80} & \textbf{97.00} & 41.80 & \textbf{18.40} & \textbf{98.20} & \textbf{99.40} & \textbf{99.52} & 22.60 & \textbf{98.13} & \textbf{72.80} & \textbf{48.60} & \textbf{76.56} \\
\bottomrule\end{tabular}
\endgroup
\par\vspace{3pt}\begin{minipage}{\linewidth}\fontsize{8}{9.3}\selectfont Task abbreviations follow Table~\ref{tab:exp-ruler-tasks}.\end{minipage}
\end{table}

\begin{table}[!htbp]
\centering
\caption{\textbf{Complete RULER scores: Qwen3-8B, 8K.} Compression ratios 50\% and 75\%.}
\label{tab:exp:app-ruler-8b-8192}
\begingroup\fontsize{7.6}{9.4}\selectfont
\setlength{\tabcolsep}{1.2pt}\renewcommand{\arraystretch}{1.18}
\setlength{\ExpMethodWidth}{0.159\linewidth}\setlength{\ExpNumWidth}{\dimexpr(\linewidth-\ExpMethodWidth-28\tabcolsep)/14\relax}\begin{tabular}{@{}p{\ExpMethodWidth}*{14}{>{\raggedleft\arraybackslash}p{\ExpNumWidth}}@{}}
\toprule
Method & S1 & S2 & S3 & MK1 & MK2 & MK3 & MV & MQ & VT & CWE & FWE & QA1 & QA2 & Avg. \\
\midrule
\rowcolor{ExpReference}
Full-KV & 100.00 & 100.00 & 100.00 & 99.80 & 99.60 & 99.60 & 100.00 & 100.00 & 100.00 & 94.24 & 91.00 & 75.80 & 60.20 & 93.86 \\
\midrule
\multicolumn{15}{l}{\textbf{Compression ratio: 50\%}} \\
SnapKV & 88.40 & 73.20 & 3.00 & 42.40 & 29.80 & 24.40 & 37.95 & 37.70 & 88.80 & 91.74 & \textbf{92.33} & 49.00 & 47.40 & 54.32 \\
KeyDiff & \textbf{100.00} & 99.80 & 99.00 & 95.20 & 53.40 & 17.00 & 99.70 & 99.55 & \textbf{100.00} & 93.20 & 87.67 & 34.60 & 26.40 & 77.35 \\
DMA & 99.00 & 82.40 & 80.80 & 83.60 & 75.60 & 1.60 & 75.60 & 89.25 & 99.16 & 77.28 & 88.93 & 74.00 & 50.60 & 75.22 \\
TrimKV & \textbf{100.00} & \textbf{100.00} & 99.80 & 99.40 & \textbf{99.00} & 21.00 & 99.55 & 99.90 & \textbf{100.00} & \textbf{94.30} & 90.40 & 72.00 & 58.60 & 87.23 \\
KVpop & \textbf{100.00} & \textbf{100.00} & 99.60 & 98.00 & 96.80 & 48.20 & 97.95 & 99.45 & \textbf{100.00} & 93.00 & 88.53 & 67.60 & 55.20 & 88.03 \\
\rowcolor{ExpOurs}
\textbf{IndexMem++} & \textbf{100.00} & \textbf{100.00} & \textbf{100.00} & \textbf{100.00} & \textbf{99.00} & \textbf{99.20} & \textbf{100.00} & \textbf{100.00} & \textbf{100.00} & 94.10 & 90.80 & \textbf{75.00} & \textbf{61.20} & \textbf{93.79} \\
\midrule
\multicolumn{15}{l}{\textbf{Compression ratio: 75\%}} \\
SnapKV & 76.00 & 21.20 & 2.40 & 16.00 & 5.80 & 4.40 & 10.70 & 12.40 & 66.00 & 78.32 & 89.33 & 36.20 & 38.00 & 35.13 \\
KeyDiff & \textbf{100.00} & 95.20 & 94.80 & 86.60 & 6.80 & 1.00 & 98.20 & 95.90 & \textbf{100.00} & 83.62 & 74.87 & 20.20 & 15.60 & 67.14 \\
DMA & 91.20 & 26.00 & 13.00 & 34.40 & 8.20 & 0.00 & 18.85 & 29.05 & 92.72 & 58.24 & 80.67 & 55.00 & 35.60 & 41.76 \\
TrimKV & \textbf{100.00} & \textbf{100.00} & \textbf{99.60} & 97.80 & 51.00 & 4.40 & 97.80 & 99.35 & \textbf{100.00} & 91.34 & 91.27 & 58.00 & 52.80 & 80.26 \\
KVpop & \textbf{100.00} & 94.20 & 83.20 & 79.20 & 37.80 & 2.00 & 89.55 & 89.70 & \textbf{100.00} & \textbf{92.06} & 86.93 & 44.20 & 46.40 & 72.71 \\
\rowcolor{ExpOurs}
\textbf{IndexMem++} & \textbf{100.00} & \textbf{100.00} & 96.40 & \textbf{99.20} & \textbf{97.80} & \textbf{73.00} & \textbf{98.80} & \textbf{100.00} & \textbf{100.00} & 83.18 & \textbf{91.40} & \textbf{71.40} & \textbf{60.60} & \textbf{90.14} \\
\bottomrule\end{tabular}
\endgroup
\par\vspace{3pt}\begin{minipage}{\linewidth}\fontsize{8}{9.3}\selectfont Task abbreviations follow Table~\ref{tab:exp-ruler-tasks}.\end{minipage}
\end{table}

\begin{table}[!htbp]
\centering
\caption{\textbf{Complete RULER scores: Qwen3-8B, 16K.} Compression ratios 50\% and 75\%.}
\label{tab:exp:app-ruler-8b-16384}
\begingroup\fontsize{7.6}{9.4}\selectfont
\setlength{\tabcolsep}{1.2pt}\renewcommand{\arraystretch}{1.18}
\setlength{\ExpMethodWidth}{0.159\linewidth}\setlength{\ExpNumWidth}{\dimexpr(\linewidth-\ExpMethodWidth-28\tabcolsep)/14\relax}\begin{tabular}{@{}p{\ExpMethodWidth}*{14}{>{\raggedleft\arraybackslash}p{\ExpNumWidth}}@{}}
\toprule
Method & S1 & S2 & S3 & MK1 & MK2 & MK3 & MV & MQ & VT & CWE & FWE & QA1 & QA2 & Avg. \\
\midrule
\rowcolor{ExpReference}
Full-KV & 100.00 & 100.00 & 99.80 & 99.60 & 100.00 & 99.60 & 99.60 & 100.00 & 100.00 & 82.54 & 93.80 & 75.20 & 58.40 & 92.96 \\
\midrule
\multicolumn{15}{l}{\textbf{Compression ratio: 50\%}} \\
SnapKV & 98.20 & 94.80 & 9.20 & 67.60 & 33.40 & 18.40 & 63.80 & 67.95 & 99.00 & 77.86 & 95.60 & 44.60 & 45.20 & 62.74 \\
KeyDiff & \textbf{100.00} & \textbf{100.00} & 99.00 & 82.20 & 31.60 & 13.60 & 98.70 & 98.00 & \textbf{100.00} & \textbf{86.40} & 94.67 & 33.00 & 30.60 & 74.44 \\
DMA & 99.60 & 80.40 & 79.80 & 79.40 & 61.80 & 0.80 & 73.95 & 90.25 & 98.56 & 55.06 & 94.47 & 68.40 & 50.60 & 71.78 \\
TrimKV & \textbf{100.00} & \textbf{100.00} & 99.60 & 97.80 & 98.80 & 17.80 & 97.65 & 99.70 & \textbf{100.00} & 82.04 & \textbf{97.07} & 70.40 & 56.00 & 85.91 \\
KVpop & \textbf{100.00} & \textbf{100.00} & 99.60 & 98.80 & 84.80 & 40.00 & \textbf{99.30} & 99.25 & \textbf{100.00} & 74.40 & 92.47 & 70.20 & 52.60 & 85.49 \\
\rowcolor{ExpOurs}
\textbf{IndexMem++} & \textbf{100.00} & \textbf{100.00} & \textbf{99.80} & \textbf{99.80} & \textbf{99.40} & \textbf{79.40} & \textbf{99.30} & \textbf{99.90} & \textbf{100.00} & 76.30 & 95.40 & \textbf{75.20} & \textbf{56.80} & \textbf{90.87} \\
\midrule
\multicolumn{15}{l}{\textbf{Compression ratio: 75\%}} \\
SnapKV & 91.40 & 58.60 & 2.40 & 24.80 & 7.80 & 3.80 & 17.10 & 20.45 & 94.96 & 58.02 & 93.87 & 33.00 & 33.80 & 41.54 \\
KeyDiff & \textbf{100.00} & 97.00 & 95.40 & 70.60 & 5.00 & 0.80 & 96.95 & 94.35 & \textbf{100.00} & \textbf{82.92} & 89.13 & 19.80 & 17.40 & 66.87 \\
DMA & 94.80 & 25.00 & 7.80 & 20.00 & 2.40 & 0.00 & 13.05 & 15.90 & 90.48 & 34.20 & 83.00 & 48.20 & 35.60 & 36.19 \\
TrimKV & \textbf{100.00} & \textbf{100.00} & \textbf{99.20} & 96.40 & 45.60 & 4.60 & 95.80 & 98.50 & \textbf{100.00} & 77.66 & \textbf{98.27} & 62.20 & 50.00 & 79.09 \\
KVpop & \textbf{100.00} & 93.20 & 80.20 & 83.40 & 31.80 & 1.60 & 91.35 & 89.95 & \textbf{100.00} & 67.84 & 88.13 & 63.80 & 42.20 & 71.81 \\
\rowcolor{ExpOurs}
\textbf{IndexMem++} & \textbf{100.00} & \textbf{100.00} & 98.40 & \textbf{98.80} & \textbf{51.20} & \textbf{22.40} & \textbf{98.95} & \textbf{99.70} & 99.96 & 56.04 & 96.47 & \textbf{73.80} & \textbf{56.40} & \textbf{80.93} \\
\bottomrule\end{tabular}
\endgroup
\par\vspace{3pt}\begin{minipage}{\linewidth}\fontsize{8}{9.3}\selectfont Task abbreviations follow Table~\ref{tab:exp-ruler-tasks}.\end{minipage}
\end{table}

\subsection{Complete LongBench results}
\label{app:exp:longbench}
Table~\ref{tab:exp:app-longbench} reports the same \ExpLongBenchTaskCount{} tasks and methods as Table~\ref{tab:exp-longbench}, with \ExpLongBenchSampleCount{} examples in the displayed task set. Panel (a) contains document QA and summarization; panel (b) contains the remaining tasks and the overall average. The budget accounting is given below.
\begin{table}[!htbp]
\centering
\caption{\textbf{Complete LongBench results on Qwen3-8B.} 15 tasks (3,550 examples), at compression ratios 50\% and 75\%.}
\label{tab:exp:app-longbench}
\begingroup\fontsize{8.5}{10.3}\selectfont
\setlength{\tabcolsep}{2pt}\renewcommand{\arraystretch}{1.15}
\noindent\textbf{(a) Single-document QA, multi-document QA, and summarization}\par\vspace{4pt}
\setlength{\ExpMethodWidth}{0.17\linewidth}\setlength{\ExpNumWidth}{\dimexpr(\linewidth-\ExpMethodWidth-18\tabcolsep)/9\relax}\begin{tabular}{@{}p{\ExpMethodWidth}*{9}{>{\raggedleft\arraybackslash}p{\ExpNumWidth}}@{}}
\toprule
Method & NQA & Qasper & MFQA & Hotpot & 2Wiki & MuSiQue & GovR & QMSum & MNews \\
\midrule
\rowcolor{ExpReference}
Full-KV & 26.31 & 43.98 & 55.59 & 62.30 & 48.22 & 35.23 & 33.55 & 24.50 & 24.77 \\
\midrule
\multicolumn{10}{l}{\textbf{Compression ratio: 50\%}} \\
SnapKV & 23.50 & 37.36 & 45.94 & 59.48 & 47.17 & 32.91 & 31.91 & 22.63 & 23.26 \\
KeyDiff & 16.81 & 29.46 & 40.33 & 36.23 & 34.46 & 16.03 & 29.34 & 22.37 & 21.66 \\
DMA & 22.40 & 42.19 & 52.01 & 55.56 & 39.37 & 33.24 & 32.21 & 23.22 & 24.16 \\
TrimKV & \textbf{26.25} & \textbf{47.66} & 51.89 & 59.33 & 43.08 & \textbf{36.45} & 33.19 & 24.09 & 24.31 \\
\rowcolor{ExpOurs}
\textbf{IndexMem++} & 25.87 & 43.38 & \textbf{53.20} & \textbf{62.84} & \textbf{48.82} & 35.51 & \textbf{33.64} & \textbf{24.50} & \textbf{24.79} \\
\midrule
\multicolumn{10}{l}{\textbf{Compression ratio: 75\%}} \\
SnapKV & 22.09 & 28.96 & 34.71 & 52.64 & 39.71 & 26.71 & 29.52 & 21.37 & 21.46 \\
KeyDiff & 12.11 & 19.56 & 28.76 & 20.70 & 27.82 & 9.54 & 21.91 & 21.52 & 13.94 \\
DMA & 17.30 & 33.19 & 37.01 & 40.60 & 38.11 & 24.45 & 28.05 & 21.26 & 21.31 \\
TrimKV & \textbf{24.75} & \textbf{44.34} & 49.82 & 59.53 & 42.40 & 31.81 & \textbf{32.87} & \textbf{23.72} & \textbf{24.40} \\
\rowcolor{ExpOurs}
\textbf{IndexMem++} & 24.66 & 40.61 & \textbf{50.81} & \textbf{61.57} & \textbf{44.82} & \textbf{35.89} & 32.07 & 23.25 & 23.35 \\
\bottomrule\end{tabular}
\par\vspace{9pt}
\noindent\textbf{(b) Few-shot, synthetic, and code tasks; overall average}\par\vspace{4pt}
\setlength{\ExpMethodWidth}{0.17\linewidth}\setlength{\ExpNumWidth}{\dimexpr(\linewidth-\ExpMethodWidth-14\tabcolsep)/7\relax}\begin{tabular}{@{}p{\ExpMethodWidth}*{7}{>{\raggedleft\arraybackslash}p{\ExpNumWidth}}@{}}
\toprule
Method & TriviaQA & SAMSum & PCount & PRetr. & LCC & Repo-P & Avg. \\
\midrule
\rowcolor{ExpReference}
Full-KV & 90.56 & 40.23 & 10.00 & 90.60 & 64.74 & 60.20 & 47.39 \\
\midrule
\multicolumn{8}{l}{\textbf{Compression ratio: 50\%}} \\
SnapKV & 89.74 & 40.77 & 9.00 & 91.00 & \textbf{66.20} & \textbf{60.00} & 45.39 \\
KeyDiff & 85.86 & 38.68 & 4.89 & 86.33 & 40.74 & 50.47 & 36.91 \\
DMA & 85.16 & 43.50 & 3.50 & 67.00 & 57.45 & 52.80 & 42.25 \\
TrimKV & \textbf{90.85} & \textbf{44.76} & 3.50 & \textbf{100.00} & 57.13 & 55.56 & 46.54 \\
\rowcolor{ExpOurs}
\textbf{IndexMem++} & 90.29 & 40.25 & \textbf{10.00} & 95.22 & 65.05 & 59.59 & \textbf{47.53} \\
\midrule
\multicolumn{8}{l}{\textbf{Compression ratio: 75\%}} \\
SnapKV & \textbf{89.59} & 40.49 & \textbf{7.55} & 88.29 & \textbf{65.78} & \textbf{61.46} & 42.02 \\
KeyDiff & 81.63 & 36.19 & 4.82 & 48.33 & 25.63 & 50.43 & 28.19 \\
DMA & 77.46 & 41.91 & 1.50 & 8.50 & 53.87 & 46.30 & 32.72 \\
TrimKV & 89.42 & \textbf{43.47} & 3.00 & \textbf{100.00} & 56.50 & 55.17 & 45.41 \\
\rowcolor{ExpOurs}
\textbf{IndexMem++} & 89.50 & 39.44 & 7.50 & 94.42 & 63.79 & 58.72 & \textbf{46.03} \\
\bottomrule\end{tabular}
\endgroup
\par\vspace{3pt}\begin{minipage}{\linewidth}\fontsize{8}{9.3}\selectfont NQA = NarrativeQA; MFQA = MultiFieldQA-en; Hotpot = HotpotQA; 2Wiki = 2WikiMQA; GovR = GovReport; MNews = MultiNews; PCount = PassageCount; PRetr. = PassageRetrieval-en; Repo-P = RepoBench-P. Avg. is the equal-weight mean of all 15 displayed tasks. Budget accounting and protected regions are given in Appendix~\ref{app:exp:longcontext}.\end{minipage}
\end{table}

\paragraph{RULER budget accounting.}
The input is processed as a context followed by a question and generation suffix. RULER is evaluated at nominal context lengths of 8K and 16K. Let $N$ denote the actual tokenized context length and $c$ the compression ratio. SnapKV and KeyDiff keep $\lfloor(1-c)N\rfloor$ entries per KV head. TrimKV sets its memory size to the same fraction $(1-c)$ of each input's context length, and KVpop applies the same compression ratio through its released model variant. DMA sets its top-$K$ from the nominal length, i.e., $(1-c)$ times 8K or 16K; it keeps the full KV cache and restricts attention to the selected entries. For IndexMem++, $K$ is the average per-KV-head budget in Table~\ref{tab:exp:inference-config}. All budgets are inclusive: the attention sinks, local window, and any auxiliary slots a method maintains are funded within the budget rather than added on top. Table~\ref{tab:exp:protocol} lists the retention score and the protected region of each method.

For IndexMem++, $K$ is the average per-KV-head budget. The mass allocator redistributes the fixed layer-wide total $\sum_h B_h=H_{\mathrm{KV}}K$ across KV heads, so an individual head may hold more or fewer than $K$ entries while the layer total is preserved.

\begin{table}[!htbp]
\centering
\caption{\textbf{RULER retention criteria and protected regions.} Budgets follow the accounting above; the protected region of each method is funded within its budget rather than added on top.}
\label{tab:exp:protocol}
\begingroup\fontsize{8.7}{10.3}\selectfont
\setlength{\tabcolsep}{4pt}\renewcommand{\arraystretch}{1.18}
\begin{tabularx}{\linewidth}{@{}lXX@{}}
\toprule
Method & Retention score & Protected region (within budget) \\
\midrule
SnapKV & Observed-attention mean with width-5 average pooling & Final 64 context positions \\
KeyDiff & Cosine-based post-RoPE key scoring & Final context position \\
DMA & Query-independent value-derived gate score & Sink 4, local 128 \\
TrimKV & Released gate deltas, magnitude and lifetime & Sink 4, local 128 \\
KVpop & Recurrent scorer state & Sink 4, local 256 \\
\rowcolor{ExpOurs}
IndexMem++ & Learned gate on $h_j$; mass allocation across heads & Sink 4, local 128, CMP 64 \\
\bottomrule
\end{tabularx}
\endgroup
\end{table}

\paragraph{LongBench.}
All methods use a maximum context length of 32,768 tokens and the task-specific generation limits defined by the benchmark. Prompts longer than this limit are truncated before prefill. For compression ratio $c$, the average per-KV-head budget is $K=\lfloor(1-c)N\rfloor$, where $N$ is the prompt length after truncation; the budget is fixed once at the end of prefill and held during decoding. As on RULER, the budget is inclusive of each method's protected region, and the retention criteria and protected regions are those of Table~\ref{tab:exp:protocol}.

\paragraph{Averaging.}
RULER and LongBench scores are shown on the benchmark's 0--100 scale. RULER uses the arithmetic mean of the 13 scorer-returned task scores. LongBench uses the equal-weight mean of the \ExpLongBenchTaskCount{} displayed task scores; its category sizes are 3, 3, 3, 2, 2, and 2. Few-shot averages only TriviaQA and SAMSum. We calculate category and overall means from task scores before rounding the displayed result to two decimals.

\section{Math Reasoning Protocol}
\label{app:exp:math}

All math results use Qwen3-8B with thinking enabled, temperature 0.6, and top-$p$ 0.95. pass@1 is the fraction of correct rollouts: 500 problems with two rollouts each on Math500, and the 30 problems of AIME25 I+II with eight rollouts each. The rows of Table~\ref{tab:exp-math} come from separate evaluation runs whose settings differ as follows.

\paragraph{IndexMem++ and Full-KV.} Both apply top-$k$ 20 and the same per-rollout seed scheme, so their rows are paired. They generate at most 16{,}384 new tokens on Math500 and 32{,}768 on AIME25. IndexMem++ uses the offline per-budget head table with a $K/4$ floor and no centroid slots (Appendix~\ref{app:exp:protocol}). Their prompts end with an instruction to put the final answer within \texttt{\textbackslash boxed\{\}}; the problem sets are otherwise identical to those of the other runs.

\paragraph{R-KV and TriAttention.} Both run in the Random Attention evaluation code, with prompt plus output capped at 32{,}768 tokens, top-$k$ 20, and a per-problem seed scheme, so no paired test against IndexMem++ is available. R-KV uses $\lambda=0.5$ with smoothing and a 64-token residual window. TriAttention uses $n_{\mathrm{large}}=256$ and a 64-token residual window. Because no Qwen3-8B calibration is released, its statistics are calibrated on the question text of the 530 evaluation problems, without answers (40{,}691 tokens).

\paragraph{Random Attention.} These rows are produced by the released Random Attention code. The policy keeps the prompt and a 64-token recency window and fills the rest of the budget with a uniformly random per-KV-head subset of generated tokens. Prompt plus output is capped at 32{,}768 tokens. The runs are unseeded and apply only temperature and top-$p$, without top-$k$.

\paragraph{KVzap and Fast-KVzip.} Both use their released Qwen3-8B scorers under one shared eviction rule that holds $K$ entries per KV head at every decoding step, with four sinks and a window of $\lfloor K/4 \rfloor$ entries inside the budget. Their Math500 cells use a 125-problem subsample (drawn with seed 42) with two rollouts each; their AIME25 cells use all 30 problems. They generate at most 16{,}384 new tokens on Math500 and 32{,}768 on AIME25 and use the same seed scheme and prompt as the IndexMem++ runs. Their rows are paired only with each other.

\section{Analysis of Retained Entries}
\label{app:exp:retention}

This section gives the setup and the complete results behind Section~\ref{sec:exp:retention}.

\paragraph{Setup.} All runs use Qwen3-8B in BF16 with the IndexMem++ gate of Section~\ref{sec:exp:setup} and physical eviction. The average per-KV-head budget is $K=2{,}048$, inclusive of 64 centroid slots, four sinks, and a 128-token local window, which leaves 1{,}852 selectable entries per head under uniform allocation. The control is a recency policy with the same per-layer, per-head budgets and the same centroid slots; it keeps the most recent selectable entries. The AIME25 traces are generated by Qwen3-8B with Full-KV under the decoding settings of the free-generation runs below. We use one trace longer than 8K tokens for each of 28 problems; the other two problems have no such trace. The SWE data are 25 SWE-agent trajectories on SWE-bench tasks from 8 repositories. We replay the full message history verbatim, with its role and tool-call boundaries, rather than the per-step requests, which keep only the most recent tool observations.

Each trace is examined under two protocols. A snapshot prefills the 8K-token prefix at once and commits the compressed cache. A replay feeds the trace token by token from the initial prompt and updates the physical cache at every step; when the initial prompt is short, the allocator falls back to uniform budgets. Positional statistics exclude the sinks and the local window, which leaves 8{,}060 selectable positions, and count original entries only, since centroid slots have no position.

\paragraph{Overlap with a recency window.} For each head, we compare its retained set with the most recent selectable positions of the same size and report the fraction of retained positions that fall in this recency window. The fraction is averaged over heads within a trace and then over traces. On eight traces per domain replayed to 8K tokens, the overlap is 52.50\% on AIME25 and 48.35\% on SWE, so about half of the retained entries lie outside the recency window. The recency policy overlaps by 100\% by construction.

\paragraph{Prediction on fixed continuations.} On the same eight traces per domain, we score the 256 assistant tokens that follow the 8K-token prefix; tool and user tokens are fed to the model but not scored. Table~\ref{tab:exp:retention-nll} reports the NLL and the KL divergence from the Full-KV next-token distribution, $\mathrm{KL}(p_{\mathrm{full}}\,\|\,p)$, averaged over traces. IndexMem++ is lower than the recency policy on both metrics for all eight traces in each domain.

\begin{table}[!htbp]
\centering
\caption{Prediction on fixed continuations with Qwen3-8B at $K=2{,}048$, averaged over eight traces per domain. Lower is better.}
\label{tab:exp:retention-nll}
\begin{tabular}{lccccc}
\toprule
 & \multicolumn{3}{c}{NLL} & \multicolumn{2}{c}{KL from Full-KV} \\
\cmidrule(lr){2-4}\cmidrule(lr){5-6}
 & Full-KV & IndexMem++ & Recency & IndexMem++ & Recency \\
\midrule
AIME25 & 0.197 & 0.207 & 0.279 & 0.021 & 0.082 \\
SWE & 1.068 & 1.221 & 1.551 & 0.120 & 0.630 \\
\bottomrule
\end{tabular}
\end{table}

\paragraph{Early facts in SWE trajectories.} In each SWE trajectory, we identify facts stated early in the issue and their explicit applications later in the trajectory, between 4K and 8K tokens. We then check whether the corresponding original entries are still cached when each fact is applied. A fact counts as retained when every token of its core span has a copy in at least one layer and head. In 17 of the 25 trajectories, covering 7 of the 8 repositories, at least one fact is applied later. All 39 applied facts are retained before they are applied; 30 of them are complete within the heads of a single layer and 16 within a single head. The recency policy keeps no original entry for 37 of the 39. Facts with no observed later application are also retained: 111 of 111 at 4K tokens and 109 of 111 at 8K tokens. Retention is therefore not specific to facts that are applied later. In addition, 23 of the applied facts reappear in full later in the trajectory, for example when they are restated or re-read by a tool.

\paragraph{Layers.} We measure the similarity of two layers by the weighted Jaccard index of their retained positions, weighting each position by the number of heads in the layer that keep it. In SWE replay, adjacent layers score 0.512 and layers at least eight apart score 0.413. Adjacent layers are more similar than distant ones in all 25 SWE and all 28 AIME25 traces. A random selection that matches each head's capacity and its counts in eight coarse age bins gives 0.513 for adjacent layers, averaged over 100 draws. Taking unions, the eight heads of one layer hold original entries for 57.2\% of the selectable positions on average, four layers for 85.0\% in expectation over layer subsets, and all 36 layers for 99.9\%. The matched random selection reaches 97.7\% with four layers.

\paragraph{Heads.} Within a layer, the Jaccard index between the retained sets of two heads averages 0.372 in SWE replay, against 0.212 for the matched random selection. In 8K snapshots, the two heads with the largest budgets in each layer hold 50.2\% of the selectable capacity on average (95\% interval [49.2\%, 51.3\%]), whereas uniform allocation would give 25\%.

\paragraph{Free generation on AIME25.} For each of the 30 AIME25 problems, we generate eight samples with thinking enabled, temperature 0.6, top-$p$ 0.95, top-$k$ 20, and at most 32{,}768 new tokens. Under the configuration above, IndexMem++ reaches a pass@1 of 0.592 (142 of 240 samples correct), the recency policy 0.121 (29 of 240), and Full-KV 0.654 (157 of 240). Paired over problems, IndexMem++ exceeds the recency policy by 47.08 points (95\% bootstrap interval [33.75, 60.83]) and trails Full-KV by 6.25 points ([1.67, 11.25]).

\end{document}